\documentclass[letterpaper]{article} 
\usepackage{aaai2026}  
\usepackage{times}  
\usepackage{helvet}  
\usepackage{courier}  
\usepackage[hyphens]{url}  
\usepackage{graphicx} 
\usepackage{natbib}  
\usepackage{caption} 
\usepackage{algorithm}
\usepackage{algorithmic}

\usepackage{newfloat}
\usepackage{listings}
\DeclareCaptionStyle{ruled}{labelfont=normalfont,labelsep=colon,strut=off} 
\floatstyle{ruled}
\newfloat{listing}{tb}{lst}{}
\floatname{listing}{Listing}
\usepackage{booktabs}
\usepackage{adjustbox}
\usepackage{multirow}
\usepackage[nointegrals]{wasysym}
\usepackage{amssymb}
\usepackage{amsmath}
\usepackage{tikz}
\usetikzlibrary{positioning}
\usepackage{enumitem}
\usepackage{scalerel}
\usepackage{subfig}
\usepackage{longtable}

\DeclareUnicodeCharacter{266B}{\twonotes}

\title{SPARE: Single-Pass Annotation with Reference-Guided Evaluation for Automatic Process Supervision and Reward Modelling}
\author{
    Md Imbesat Hassan Rizvi\textsuperscript{\rm 1}, Xiaodan Zhu\textsuperscript{\rm 2}, Iryna Gurevych\textsuperscript{\rm 1}
}
\affiliations{
    \textsuperscript{\rm 1}Ubiquitous Knowledge Processing Lab (UKP Lab), Department of Computer Science and \\
    Hessian Center for AI (hessian.AI), Technical University of Darmstadt, Germany \\
    \textsuperscript{\rm 2}Department of Electrical and Computer Engineering \& Ingenuity Labs Research Institute,  \\
    Queen’s University, Canada \\

    www.ukp.tu-darmstadt.de, xiaodan.zhu@queensu.ca
}

\begin{document}

\maketitle

\begin{abstract}
Process or step-wise supervision has played a crucial role in advancing complex multi-step reasoning capabilities of Large Language Models (LLMs). However, efficient, high-quality automated process annotation remains a significant challenge. 
To address this, 
we introduce \textbf{S}ingle-\textbf{P}ass \textbf{A}nnotation with \textbf{R}eference-Guided \textbf{E}valuation (\textbf{\texttt{SPARE}}), a novel structured framework that enables efficient per-step annotation by jointly aligning solution steps to reference solutions and determine its accuracy with explicit reasoning in single generation.
We demonstrate \texttt{SPARE}'s effectiveness across four diverse datasets spanning mathematical reasoning (GSM8K, MATH), multi-hop question answering (MuSiQue-Ans), and spatial reasoning (\texttt{SpaRP}), showing consistent improvements in two applications: (1) training Process Reward Models (PRMs) for ranking and aggregating multiple generations, and (2) fine-tuning models via offline reinforcement learning for greedy decoding. On \textsc{ProcessBench}, \texttt{SPARE} demonstrates data-efficient out-of-distribution generalization, using only $\sim$16\% of training samples compared to human-labeled and other synthetically trained baselines. Additionally, it achieves competitive performance with MCTS-based methods while offering 2.3$\times$ speedup in terms of total token count. Manual analysis reveals complementary precision-recall characteristics with MCTS approaches, suggesting potential for ensemble methods. These results establish \texttt{SPARE} as a practical and scalable solution for automatic process supervision in LLM reasoning.
\end{abstract}

\begin{links}
    \link{Process Reward Models (SPARE-PRM)}{https://huggingface.co/collections/UKPLab/spare-prm}
    \link{Code}{https://github.com/UKPLab/aaai2026-spare-prm}
    \link{Extended version with Appendices}{https://www.arxiv.org/abs/2506.15498}
\end{links}

\section{Introduction}
\label{sec:introduction}

While large language models (LLMs) have demonstrated strong performance across a broad range of tasks~\cite{brown2020gpt3,wei2022emergent, wei2022cot,chowdhery2023palm,touvron2023llamaopenefficientfoundation,srivastava2023beyond}, complex multi-step reasoning still remains a challenge for LLMs even when they are trained and finetuned with ground-truth chains of thoughts~\cite{azerbayev2024llemma, yu2024metamath}. 
Self-consistency
can improve performance by voting over multiple generations, only if the answers are correct in majority of them. To address this, reward models trained to assess output correctness have gained popularity. Outcome Reward Models (ORMs)~\cite{cobbe2021trainingverifierssolvemath, yu-etal-2024-ovm} are trained using outcome supervision relying on the correctness of the final answer, while Process Reward Models (PRMs)~\cite{uesato2022solvingmathwordproblems, lightman2024lets} use process supervision that relies on the correctness of individual reasoning steps. 

PRMs achieve better performance due to the targeted step-level feedback but suffer from expensive and complex annotation requirements. Human-supervision
~\cite{uesato2022solvingmathwordproblems, lightman2024lets} is very demanding in terms of highly skilled human evaluators, motivating efforts toward automatic process annotation largely driven by Monte Carlo Tree Search (MCTS)-based methods~\cite{wang-etal-2024-math, wang-etal-2024-multi-step, luo2024improvemathematicalreasoninglanguage, zhang2024restmcts}. In MCTS-based approaches, models are initially trained on ground-truth reasoning traces and answers through supervised fine-tuning. However, during step evaluation, these methods overlook the valuable step-by-step information \textit{already present} in the reference ground-truth rationales. Instead, they rely exclusively on final answer matching across multiple model rollouts, resulting in both computational inefficiency and \textit{under-utilization} of the data already available at hand.

\begin{figure*}
    \centering
    \includegraphics[width=\linewidth]{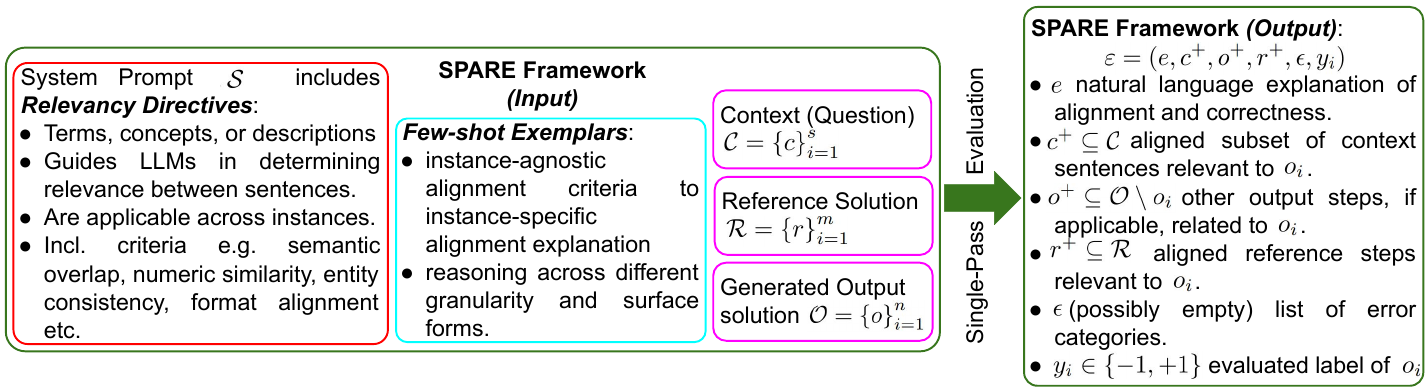}
    \caption{We propose a unified, single-stage framework: \texttt{\textbf{S}}ingle \texttt{\textbf{P}}ass \texttt{\textbf{A}}nnotation with \texttt{\textbf{R}}eference-Guided \texttt{\textbf{E}}valuation (\textbf{\texttt{SPARE}}: $(\mathcal{S, C, R, O}) \rightarrow \mathcal{E}$). \texttt{SPARE} produces an explanation-based step-by-step evaluation $\mathcal{E}$ of a candidate model output $\mathcal{O}$, grounded to a given context $\mathcal{C}$, reference reasoning $\mathcal{R}$, and system prompt $\mathcal{S}$ (Section~\ref{sec:SPARE}).}
    \label{fig:SPARE_overview}
\end{figure*}

Parallel efforts aim to leverage valuable signals from reference reasoning traces, that are either \textit{existing} ground-truth or \textit{synthetically} generated rationales. For instance, \citet{li-etal-2023-making, khalifa-etal-2023-grace} generated step-level annotations by decomposing candidate and reference solutions into individual steps, performing alignment using dataset-specific discriminative models, 
and annotating steps in a restricted context where a candidate step is matched to a single reference step. AutoPRM~\cite{chen-etal-2024-autoprm} decomposes reference solutions into sub-questions and corresponding solutions to enable process supervision. However, their approach relies on an auxiliary model for data collection, which is trained using outputs from a more capable language model. More recently, GenRM~\cite{zhang2025genrm} employed reference-guided grading to train verifiers using synthetically generated rationales as references. However, GenRM is not a process supervision (PRM) model and relies on rationales from a more capable model than the one trained as the reward model. More recently, ThinkPRM~\cite{khalifa2025thinkprm} and R-PRM~\cite{she2025rprmreasoningdrivenprocessreward} utilized a more capable model to generate synthetic verification rationales, which were subsequently filtered to retain only those whose step annotations aligned with human-labeled steps in the PRM800K dataset~\cite{lightman2024lets}. While these approaches maximize the utility of available data, they depend on step labels for initialization or filtering, labels that are the end goal of process supervision itself and may not exist for other datasets or domains.

To address these gaps, we propose \textbf{S}ingle-\textbf{P}ass \textbf{A}nnotation with \textbf{R}eference-Guided \textbf{E}valuation (\texttt{SPARE}), a general-purpose framework
that enables automatic process supervision through step-level evaluation of model responses by leveraging intermediate steps from reference reasoning traces. \texttt{SPARE} strikes a balance between leveraging available data and maintaining broad applicability across domains, using only a single model throughout the process.  Concretely, it introduces a generic, structured evaluation scheme that (i) emphasizes explicit reasoning during step evaluation, and (ii) supports multi-step alignment between model outputs and references. It accomplishes this by encoding instance-agnostic alignment and evaluation criteria in the system prompt, complemented by in-context exemplars that illustrate how to apply these guidelines to specific instances (Figure~\ref{fig:SPARE_overview}). This design enables single-pass evaluation with additive scaling relative to the token lengths of the response and reference. 
In summary, our contributions are:

\begin{itemize}
\item We introduce \texttt{SPARE}, a general, single-pass, and structured reference-guided evaluation framework for process annotation, which emphasizes explicit reasoning and multi-step alignment. Notably, \texttt{SPARE} is agnostic to the source of reference solutions, assuming they are of high quality. In this work, we reuse existing reference solutions from standard supervised fine-tuning (SFT) datasets without requiring additional reasoning traces.

\item We utilize \texttt{SPARE} annotations to improve LLM reasoning via: (i) training Reward Models (RMs) for ranking and aggregating multiple generations, and (ii) fine-tuning models in an offline reinforcement learning (RL) setup for greedy-decoding during inference.

\item We evaluate \texttt{SPARE} across four benchmarks—GSM8K, MATH, 
MuSiQue-Ans, 
and \texttt{SpaRP}—demonstrating consistent improvements over baselines, and out-of-distribution generalization on PROCESSBENCH with high data efficiency. It substantially reduces annotation cost versus tree-search methods, with manual analysis revealing complementary precision-recall characteristics. 
\end{itemize}

\section{Related Work}
\label{sec:related_work}

\paragraph{Reasoning abilities of LLMs.} Reasoning remains a challenging area for the Large Language Models (LLMs). Various prompting techniques, such as chain-of-thought, few-shot prompting and their variants~\cite{wei2022cot, kojima2022zeroshotreasoners, yao2023tree, hao-etal-2023-reasoning}
elicited reasoning capabilities in LLMs. Importance of individual steps while prompting~\cite{fu2023complexitybased,zhou2023leasttomost} was soon found to be crucial in successfully solving multi-step reasoning problems. While prompt-only techniques show promising results, their performances are constrained by and sensitive to prompt design and nature of tasks~\cite{ye2022unreliableprompting}. Consequently, explicitly finetuning with high-quality reasoning traces for improving LLM reasoning capabilities has become popular~\cite{yu2024metamath, luo2025wizardmathempoweringmathematicalreasoning}. 

\paragraph{Outcome and Process Supervision.} Supervised finetuning quickly results in saturation, leading to the search for other advanced techniques and better supervision signals. Outcome supervision~\cite{cobbe2021trainingverifierssolvemath, yu-etal-2024-ovm} relies on signal based on the final answer, and hence, is easier to obtain. Process supervision offers advantages in the form of fine-grained feedback from individual reasoning steps, however, early work~\cite{uesato2022solvingmathwordproblems, pan2023reinforcestepbystep, lightman2024lets} relied on time-consuming and costly human annotation. To alleviate this problem, several recent approaches have emerged for automating process supervision. Monte-Carlo Tree Search (MCTS) based approaches~\cite{wang-etal-2024-math, wang-etal-2024-multi-step, luo2024improvemathematicalreasoninglanguage, zhang2024restmcts} target obtaining process annotation by several continuations from intermediate steps whose correctness are evaluated based on the final step. Parallel work has explored leveraging reference reasoning traces, either ground-truth or synthetic, for supervision. Prior approaches decompose solutions into steps for alignment~\cite{li-etal-2023-making, khalifa-etal-2023-grace}, often relying on dataset-specific models with limited generalization. Others, like AutoPRM~\cite{chen-etal-2024-autoprm}, use sub-question decomposition but depend on auxiliary models trained with outputs from stronger LLMs. GenRM~\cite{zhang2025genrm}, ThinkPRM~\cite{khalifa2025thinkprm}, and R-PRM~\cite{she2025rprmreasoningdrivenprocessreward} use synthetic rationales from more capable models for training verifiers, but do not provide a general-purpose, reference-guided process supervision framework. In contrast to these efforts, our work proposes a unified, single-pass, and structured evaluation framework for automatic process annotation, enabling flexible alignment and multi-step comparison with reference solutions. We further demonstrate its effectiveness across both fine-tuning and verification settings.


\section{Our Approach}
\label{sec:methodology}

\subsection{Single-Pass Annotation with Reference-Guided Evaluation (SPARE)}
\label{sec:SPARE}

We propose \textbf{S}ingle-\textbf{P}ass \textbf{A}nnotation with \textbf{R}eference-Guided \textbf{E}valuation (\textbf{\texttt{SPARE}}) as a unified LLM-driven framework for fine-grained evaluation of model-generated reasoning steps with respect to a given context and reference solution. \texttt{SPARE} \textit{jointly} infers with explicit reasoning (i) the alignment of each output step with relevant context and reference steps, and (ii) its correctness label. Concretely, given:

\begin{itemize}
    \item a reference reasoning path $\mathcal{R}={\{r\}}_{i=1}^{m}$ (with $m$ steps),
    \item a model-generated output $\mathcal{O}={\{o\}}_{i=1}^{n}$ (with $n$ steps),
    \item a contextual question $\mathcal{C}={\{c\}}_{i=1}^{s}$ (with $s$ sentences), and
    \item a system prompt $\mathcal{S}$ defining evaluation guidelines,
\end{itemize}

An answer or outcome annotation $y \in \mathbb{R}$ is a score indicating a measure of correctness of the model's output. Most commonly, $y = \mathbb{I}(o_n = r_m)$; i.e., the output's answer matches with the reference reasoning answer. In contrast, a process annotation $\mathcal{Y} = {\{y \mid y \in \mathbb{R}\}}_{i=1}^{n}$ is a sequence of scalar scores assigned to the corresponding steps $o_i$. 

We devise an evaluation sequence $\mathcal{E} = {\{\varepsilon\}}_{i=1}^{n}$ where each step $o_i$ is annotated with alignment and correctness information, such that each $\varepsilon_i$ consists of a structured tuple:

\[\varepsilon = (e,c^+,o^+,r^+,\epsilon,y_i)\]

where $e$ is a natural language explanation justifying the evaluation $y_i \in \{{-1, +1\}}$, while referring to:
\begin{itemize}
    \item $c^+ \subseteq \mathcal{C}$ is the subset of context sentences relevant to $o_i$,
    \item $o^+ \subseteq \mathcal{O} \setminus {o_i}$ contains other output steps related to $o_i$,
    \item $r^+ \subseteq \mathcal{R}$ are the reference steps relevant to $o_i$,
    \item $\epsilon$ is a (possibly empty) list of error categories.
\end{itemize} 

\begin{table*}
    \centering
    \small
    \begin{tabular}{l | c c c c c | c c c c}
        \toprule
         & \multicolumn{5}{c|}{\textbf{Aggregation / Ranking}} &  \multicolumn{4}{c}{\textbf{Model Finetuning}} \\
         \cmidrule{2-10} 
        \textbf{Dataset} & \textbf{\texttt{SC}} & \textbf{\texttt{ORM}} & \textbf{\texttt{ORM}} & \textbf{\texttt{SPARE}} & \textbf{\texttt{SPARE}} & \textbf{\texttt{SFT-1}} & \textbf{\texttt{SFT-2}} & \textbf{\texttt{SFT-1}} & \textbf{\texttt{SFT-1}} \\
        & & \textbf{\texttt{(BoN)}} & \textbf{\texttt{+ SC}} & \textbf{\texttt{(BoN)}} & \textbf{\texttt{+ SC}} & & & \textbf{\texttt{+ Out.}} & \textbf{\texttt{+ SPARE}} \\
        \midrule 
        GSM8K & 74.9 & 79.7 & 79.8 & \underline{80.0} & \textbf{80.3} & \textbf{70.4} & \textbf{70.4} & 69.0 & \underline{69.8} \\
        MATH$^\ast$ & 23.4 & 20.2 & \underline{23.8} & 20.9 & \textbf{24.1} & 21.2 & 22.1 & \underline{23.1} & \textbf{23.4} \\
        MuSiQue-Ans & 19.7 / 25.2 & 33.4 / \underline{45.4} & \underline{34.8} / 44.5 & \textbf{34.9} / \textbf{45.5} & 32.1 / 40.4 & 23.6 / 32.5 & 26.3 / 35.1 & \underline{38.2} / \underline{49.9} & \textbf{38.9} / \textbf{50.5} \\
        SpaRP-S & 25.4 / 34.4 & \underline{41.7} / \underline{49.8} & \underline{41.7} / \underline{49.8} & \textbf{43.7} / \textbf{50.0} & 39.6 / 46.9 & 23.2 / 35.0 & \underline{39.9} / 47.1 & 39.2 / \underline{49.8} & \textbf{40.1} / \textbf{51.0} \\
        \bottomrule
    \end{tabular}
    \caption{
    Llama-3 8B Instruct performance. \textbf{Bold} means best; \underline{underline} means second-best. \textbf{Aggregation}/\textbf{Ranking} on N=20 generations from Llama-3 8B SFT iteration 1. \textbf{SC} means Self-Consistency, \textbf{BoN} means Best-of-N sampling. Metrics averaged over 3 independent runs. \textbf{Finetuning} results use greedy decoding. $^\ast$ indicates BoN / SC results reported on MATH-500.
    }    \label{tab:main_results}
\end{table*}

\paragraph{Joint Alignment and Evaluation via In-Context Learning (ICL).} The core innovation of \texttt{SPARE} lies in its single-pass framework that leverages the reasoning and evaluative capabilities of large language models (LLMs) through In-Context Learning (ICL) to \textit{jointly} infer, first the step alignment and then the step correctness, in a single generation. This approach parallels Natural Language Inference (NLI) with evidence localization, i.e., not merely determining whether a hypothesis (or a key fact) is entailed by a premise (or a document), but also identifying which textual components support that entailment. Similar strategies have proven effective in fine-grained summarization evaluation~\cite{song-etal-2024-finesure}. 

We extend this paradigm of localized evidence to multi-step reasoning evaluation by enabling accurate and context-sensitive step alignment through:

\begin{itemize}
    \item \textbf{Relevancy directives} -- Terms, concepts or natural language descriptions embedded in the system prompt ($\mathcal{S}$), which guide the LLM in evaluating the relevancy between two steps, say $o_i$ and $r_j$. These directives are broadly applicable across instances and include criteria such as semantic overlap, computational or numeric similarity, entity or variable consistency, and structural or format alignment. 
    \item \textbf{Few-shot exemplars} that ground the instance-agnostic generic alignment criteria to instance-specific alignment explanation. The exemplars are created highlighting reasoning across varying granularity and surface forms, including both single- and multi-step alignment scenarios.
\end{itemize}

Conditioned on these instructions and exemplars, the LLM is prompted to jointly:

\begin{itemize}
    \item Reason and identify the aligned step(s) in the reference solution $\mathcal{R}$, context sentences $\mathcal{C}$, or peer output steps in $\mathcal{O}$ for each generated step $o_i$.
    \item Reason and explain the correctness label $y_i$ of the aligned step, optionally specifying the error type (e.g., \texttt{NUMERIC}, \texttt{COMPREHENSION}) when applicable.
\end{itemize}

\paragraph{Alignment Scenarios.} To accommodate differences in reasoning granularity ($n \neq m$), we incorporate detailed guidance in the system prompt and design few-shot exemplars that capture the following flexible alignment scenarios:

\begin{enumerate}
    \item \textit{One-to-one} -- Most simple alignment where one output step aligns directly and completely with \textit{at most} one step, making it sufficient for evaluation. The alignment can take one of the forms: (i) a single reference reasoning step ($o_i \mapsto r_j$), (ii) a single context sentence ($o_i \mapsto c_k$), (iii) follows directly from or complements another output step ($o_i \mapsto o_l$), or (iv) no alignment at all ($o_i \mapsto \varnothing$). 
    \item \textit{One-to-many} -- An output step requires alignment with \textit{at least} two steps for its evaluation. Such an alignment is \textit{necessarily} required for:
    \begin{enumerate}[label=\roman*)]
        \item \textit{Composite output steps} -- The model output step $o_i$ omits minor intermediary steps or merges multiple steps into one. Its correctness must be evaluated against multiple reference steps  $r_j$ and $c_k$. 
        \item \textit{Composite reference steps} -- The model output step $o_i$ is simple while reference steps are composite. Its correctness must be evaluated in conjunction with at least one other output step $o_l$ and at least one reference step $r_j$ or context sentence $c_k$. 
    \end{enumerate}
\end{enumerate}

In summary, our \texttt{SPARE} framework defines step correctness through LLM-mediated joint alignment and evaluation, where steps are contextualized within the broader reasoning structure through explicit reference to supporting evidence. Combined with structured explanations, \texttt{SPARE} accommodates surface form variations 
without penalizing alternate but valid solution paths, enabling step-level automatic process supervision. A complete example for the MATH dataset, including system prompt, LLM output, and example failure modes are shown in 
Appendix~A.
While LLM-based approaches may inherit model biases and errors, we mitigate these in \texttt{SPARE} through broad relevancy directives and diverse few-shot exemplars that balance correctness, reasoning diversity, and topic coverage. We note that annotation quality depends on reference solution quality; we therefore evaluate \texttt{SPARE} with existing clean references to establish its efficacy. For noisy or synthetically obtained references, we expect errors to remain localized to affected steps due to \texttt{SPARE}'s local multi-step alignment, with robustness further improvable through cross-reference consistency checks. We leave experimentation under noise, as well as extensions to multilingual and multimodal contexts, for future work.

\subsection{Training Approach}
\label{sec:training}

\subsubsection{\texttt{SPARE}--based Process Reward Model (\texttt{SPARE}--PRM)}
\label{sec:verifier_training}
We utilize the step-level evaluations $y_i$ obtained through \texttt{SPARE} as direct reward signals to train process reward models. The \texttt{SPARE}-PRM is trained in a stepwise classification setting, using the following 
cross-entropy loss:
\begin{multline}
    \mathcal{L}_{PRM} = - \sum_{i=1}^{n} \biggl( y_i~\operatorname{log}~\sigma(r_{\theta}(\mathcal{C},o_{1:i})) + \\ (1-y_i)~\operatorname{log}~(1-\sigma(r_{\theta}(\mathcal{C},o_{1:i}))) \biggr)
\end{multline}
where $o_{1:i}$ is the sub-sequence of output $\mathcal{O}$ till the $i^{th}$ step. Unlike ORMs which predict a single solution score for $\mathcal{O}$, PRMs generate a probability sequence $\mathcal{P} = \{p_i\}_{i=1}^n$ for each step $o_i \in \mathcal{O}$. 
While $\operatorname{min}$ and $\operatorname{prod}$ aggregation are commonly used~\cite{lightman2024lets, wang-etal-2024-math}, we adopt the $\operatorname{last}$ function to aggregate step-wise probabilities, following recent findings~\cite{wang-etal-2024-multi-step, zhang-etal-2025-lessons}, as it yields superior downstream performance.

\subsubsection{\texttt{SPARE}--based Finetuning (\texttt{SPARE}--ORPO)}
\label{sec:finetuning}
We propose \texttt{SPARE}-based fine-tuning to enhance model reasoning capabilities. The step-by-step process annotations $\mathcal{Y} = \{y_i\}_{i=1}^{n}$, derived using the \texttt{SPARE} framework, can be effectively integrated with both online and offline Reinforcement Learning (RL). For ease of implementation, training stability, and resource efficiency, we use Odds Ratio Preference Optimization (ORPO)
for preference training over \textit{chosen} and \textit{rejected} pairs ($\mathcal{O}_w$, $\mathcal{O}_l$).

In \texttt{SPARE}-ORPO, we compute mean step annotation $\bar{y} = \frac{1}{n} \sum y_i$ as reasoning score and combine it with answer correctness $y$ to form score tuple ($y$, $\bar{y}$) for preference pairs, where $y_w = 1$, $y_l = -1$, and $\bar{y}_w > \bar{y}_l$. Thus chosen solutions exceed rejected solutions in both reasoning quality and answer accuracy. Conversely, \texttt{Outcome}-ORPO uses preference pairs ($\mathcal{O}_w$, $\mathcal{O}_l$) based solely on answer correctness, i.e., $y_w = 1$ and $y_l = -1$.


\section{Experiment Results}
\subsection{Experimental Set-up}
\label{sec:experiments}
\paragraph{Datasets.} We conduct extensive experiments over a suite of reasoning datasets\footnote{
We use 90:10 train/dev splits for GSM8K and MATH, and 80:20 for MuSiQue-Ans, as these lack official dev-sets.
}:

\begin{itemize}
    \item \textbf{Mathematical Reasoning}. We use two mathematical datasets, GSM8K~\cite{cobbe2021trainingverifierssolvemath}, which is a collection of grade school math word problems, and MATH~\cite{hendrycks-2021-competition-math}, which contains high school competition-level math problems across seven diverse topics. Following standard practice in the verification setting, we use the MATH-500 subset~\cite{lightman2024lets} for test-time evaluation involving multiple generations.
    
    \item \textbf{Question-Answering}. We use 
    MuSiQue-Ans 
    dataset~\cite{trivedi-etal-2022-musique}, a challenging multi-hop question-answering dataset constructed by composing six diverse reasoning graphs of sub-questions from five different sources.
    
    \item \textbf{Spatial Reasoning}. We use the small \texttt{SpaRP}~\cite{rizvi-etal-2024-sparc}, i.e., \texttt{SpaRP-S} dataset, which comprises four textual spatial reasoning sub-datasets covering various spatial characterizations. \texttt{SpaRP} requires spatial relation composition to deduce relations between two objects when their direct relation is not provided in the context.
\end{itemize}

\begin{table}
    \centering
    \small
    \begin{tabular}{l c c c c}
    \toprule
    \textbf{Model} & \textbf{SC} & \textbf{ORM+SC} & \textbf{PRM+SC} & $\Delta$\textbf{RM} \\
    \midrule
    \multicolumn{5}{l}{\texttt{SPARE}-Llama3-8B:} \\
    \quad N=20 & 23.4 & 23.8 & 24.1 & 0.3 \\
    \quad N=256 & 30.5 & 31.2 & 32.4 & 1.2 \\
    \multicolumn{5}{l}{\texttt{SPARE}-Qwen2.5-3B (N=20):} \\
    \quad Qwen2.5-3B Gen. & 31.4 & 33.8 & 34.6 & 0.8 \\
    \quad Qwen2.5-3B Gen.$^\dagger$ & 66.6 & 67.6 & 68.8 & 1.2 \\
    \quad Qwen2.5-32B Gen. & 64.6 & 65.6 & 66.0 & 0.4 \\
    \midrule
    \multicolumn{5}{l}{MS-Mistral-7B:} \\
    \quad SFT Gen. & 35.1 & 38.0 & 38.3 & 0.3 \\
    \quad Process RL Gen. & 42.3 & 43.1 & 43.5 & 0.4 \\
    MS-DeepSeek-67B & 45.4 & 47.0 & 48.1 & 1.1 \\
    R-MCTS$^\ast$-Mistral-7B & 35.1 & 38.0 & 39.0 & 1.0 \\
    \bottomrule
    \end{tabular}
    \caption{
    \texttt{SPARE} performance across setups contextualized against PRMs with comparable training paradigms. N: generation count; MS: Math-Shepherd; R-MCTS$^\ast$: Rest-MCTS$^\ast$. Results for external models  from their publications at N=256 generations. Results with $^\dagger$ reported on generation length of 2048 using pre-trained model.
    }
    \label{tab:contextual_performance}
\end{table}

\begin{figure}%
    \centering
    \subfloat[\centering Level-wise Performance]{{\includegraphics[width=0.485\columnwidth]{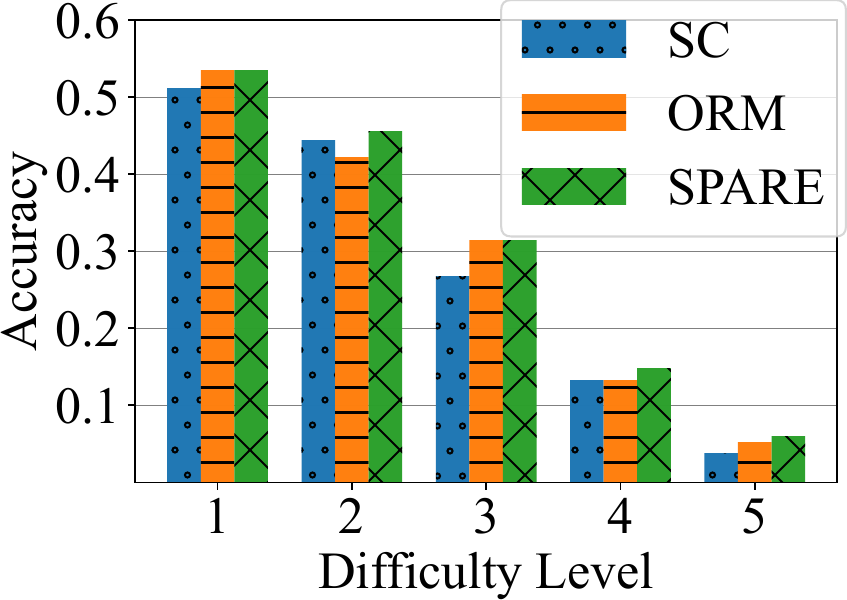}\label{subfig:level_acc} }}\hfill%
    \subfloat[\centering Inference-time Scaling]{{\includegraphics[width=0.485\columnwidth]{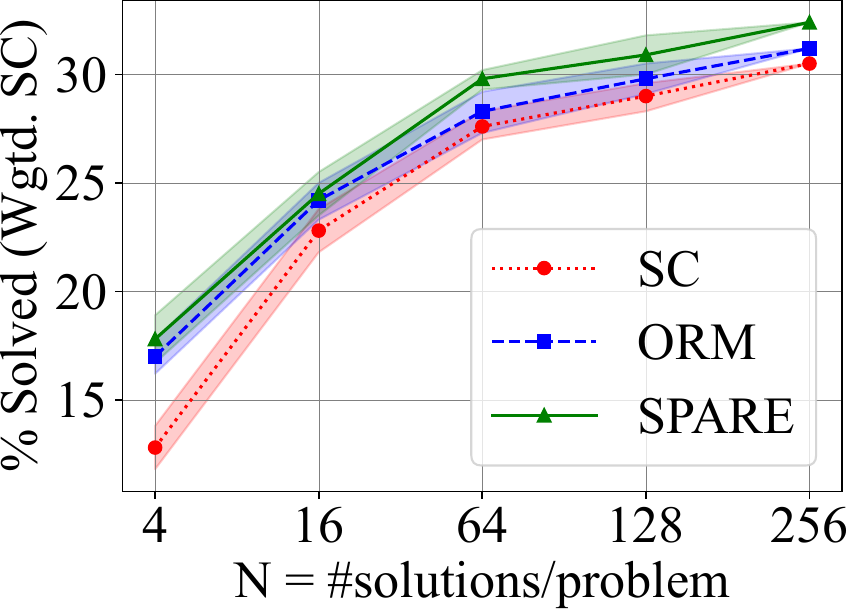}\label{subfig:inference_scaling} }}\hfill%
    \caption{
    Llama-3 8B Instruct performance across SC, ORM-weighted, and \texttt{SPARE} PRM-weighted consistency by difficulty level, and candidate scaling on MATH-500.
    }%
    \label{fig:scaling_and_level_performance}%
\end{figure}

\paragraph{Models.} We conduct our main experiments across all four datasets for both reward model training and instruct-model fine-tuning  using the LLama3-8B Instruct model.
Due to computational constraints, we limit additional experiments to selected datasets. To assess generalization across different  experimental setups and out-of-distribution datasets, we also report results using Qwen2.5 models. 

\paragraph{Metrics.} We report the accuracy\footnote{
Exact Match for GSM8K; competition math metric from \texttt{evaluate} library for MATH; Accuracy and F1 from official repositories for MuSiQue-Ans and from \texttt{scikit-learn} for SpaRP.
} for GSM8K and MATH datasets, accuracy and F1 for the 
MuSiQue-Ans
dataset, and the accuracy and macro-F1 for the \texttt{SpaRP} dataset. 

\paragraph{Parameter Setting.} All models are fine-tuned using the HuggingFace TRL library with QLoRA ($\alpha$=16, dropout = 0.1, rank $r$=64). Training is performed with an effective batch size of 32, learning rate of $1e-4$, a cosine scheduler with a warm up ratio of 0.03, and a maximum sequence length of 512, which is also used at inference unless otherwise specified. Experiments are conducted on 8 NVIDIA A100 GPUs (40 GB each). The number of training samples for the reward models are: 40,350 for GSM8K, 40,500 for MATH, 10K for MuSiQue, and 16K for \texttt{SpaRP}.

\paragraph{Implementation Details and Baselines.} We begin with a single-epoch supervised fine-tuning (SFT) on the training split. Next, for each problem in the training and dev-set, we generate $N=20$ solutions from the fine-tuned model using a temperature of 1. These solutions are then annotated using final answers for outcome supervision and the \texttt{SPARE} framework for process supervision.

We employ the same pretrained models\footnote{Performance could be further improved using specialized evaluators like Prometheus 2~\cite{kim-etal-2024-prometheus} or larger LLMs.
} for reference-guided step annotations using our \texttt{SPARE} framework. To account for problem diversity, we manually construct structured step-by-step evaluation exemplars per dataset---ranging from 6 for \texttt{SpaRP} to 56 for MATH---balanced for final answer correctness and covering all topics (MATH), sub-datasets (\texttt{SpaRP}), or reasoning graphs 
(MuSiQue-Ans). Each dataset is evaluated in a 5-shot setting, with exemplars selected randomly while ensuring both positive and negative examples, and using dataset-specific evaluation guidelines as system prompts. See 
Appendix~A 
for an example.

In the \textit{verification} scenario, we use process annotations from the \texttt{SPARE} framework to train \texttt{SPARE}-PRMs,
predicting special tokens for correct and incorrect steps as a classification objective (Section~\ref{sec:verifier_training}) at special end-of-step 
tokens. We benchmark these models against outcome reward models (ORMs) and majority-voted self-consistency.
To ensure balanced training, we randomly sample equal numbers of positive and negative examples. Evaluation metrics for both ORMs and PRMs are reported under two settings: (a) weighted aggregation (i.e., RM-weighted self-consistency) and (b) no aggregation, i.e., Best-of-N (BoN) sampling considering only the highest-scoring solution. Further training details and hyperparameters are provided in 
Appendix~B.

In the \textit{finetuning} scenario, 
we evaluate our \texttt{SPARE}-ORPO iteration trained on preference pairs formed using both outcome supervision and the mean reasoning scores of the step-by-step annotations (Section~\ref{sec:training}). We benchmark \texttt{SPARE}-ORPO against \texttt{Outcome}-ORPO and second iteration of Supervised Fine-Tuning (SFT) with an equivalent number of training instances. The training hyperparameter details are provided in 
Appendix~C.


\subsection{Results and Discussion}
\label{sec:results}

\paragraph{\texttt{SPARE} Improves Reward Model Training and Adapts to Diverse Reasoning Traces.} 
Table~\ref{tab:main_results} 
shows that \texttt{SPARE}-PRM performs the best across all four datasets, outperforming both ORMs and the majority-voted Self-Consistency (SC). The improvements of the best \texttt{SPARE}-PRM ranked or aggregated strategy over the best \texttt{Outcome} baselines are statistically significant ($p < 0.05$) under one-tailed paired $t$-test, with a maximum \textit{relative} improvement of 4.8\% accuracy on \texttt{SpaRP-S}. On the challenging MATH-500 dataset, it attains a relative improvement of 1.3\%, reaching an accuracy of 24.1\%. 
Notably, this improvement is consistent across difficulty levels and especially significant for more difficult problems (Figure~\ref{subfig:level_acc}). The performance scales with increasing number of generated solutions (Figure~\ref{subfig:inference_scaling}), consistently outerperforming the baselines. 

\begin{table*}
    \centering
    \begin{tabular}{l c c c c c c c c c c}
    \toprule
    \textbf{Model} & \textbf{\# Train} & \multicolumn{3}{c}{\textbf{MATH}} & \multicolumn{3}{c}{\textbf{Olymp.Bench}} & \multicolumn{3}{c}{\textbf{Omni-MATH}} \\
    & & Error & Correct & F1 & Error & Correct & F1 & Error & Correct & F1 \\
    \cmidrule(r){1-1} \cmidrule(r){2-2} \cmidrule(r){3-5} \cmidrule(r){6-8} \cmidrule(r){9-11}
    Math-Shepherd-7B$^\ast$ & 440K & 18.0 & 82.0 & 29.5 & \underline{15.0} & 71.1 & \textbf{24.8} & \textbf{14.2} & 73.0 & \underline{23.8} \\
    RLHFlow-Deepseek-8B$^\ast$ & 250K & \underline{21.4} & 80.0 & \underline{33.8} & 10.1 & 51.0 & 16.9 & 10.9 & 51.9 & 16.9 \\
    Skywork-7B$^\ast$ & -- & \textbf{43.8} & 62.2 & \textbf{53.6} & \textbf{17.9} & 31.9 & \underline{22.9} & \underline{14.0} & 41.9 & 21.0 \\
    \texttt{SPARE}-Llama3-8B & 40.5K & \ 6.1 & \textbf{91.6} & 11.4 & \ 3.3 & \textbf{87.6} & \ 6.4 & \ 2.8 & \underline{82.2} & \ 5.4 \\
    \texttt{SPARE}-Qwen2.5-3B & 40.5K & 16.0 & \underline{89.2} & 27.1 & 11.1 & \underline{85.0} & 19.6 & \underline{14.0} & \textbf{83.8} & \textbf{23.9} \\
    \cmidrule(r){1-1} \cmidrule(r){2-2} \cmidrule(r){3-5} \cmidrule(r){6-8} \cmidrule(r){9-11}
    Qwen-2.5-Math-7B-PRM800K (Human) & 250K & 48.0 & 90.1 & 62.6 & 35.7 & 87.3 & 50.7 & 29.8 & 86.1 & 44.3 \\
    \bottomrule
    \end{tabular}
    \caption{Fine-grained evaluation comparison of \texttt{SPARE} trained PRMs on \textsc{ProcessBench}~\cite{zheng-etal-2025-processbench}. Best values in \textbf{bold}, second best in \underline{underline}. Results marked with $^\ast$ are from the \textsc{ProcessBench} paper.}
    \label{tab:process_bench}
\end{table*}

\begin{figure}%
    \centering
    \subfloat[\centering MATH-500 (ORM)]{{\includegraphics[width=0.45\columnwidth]{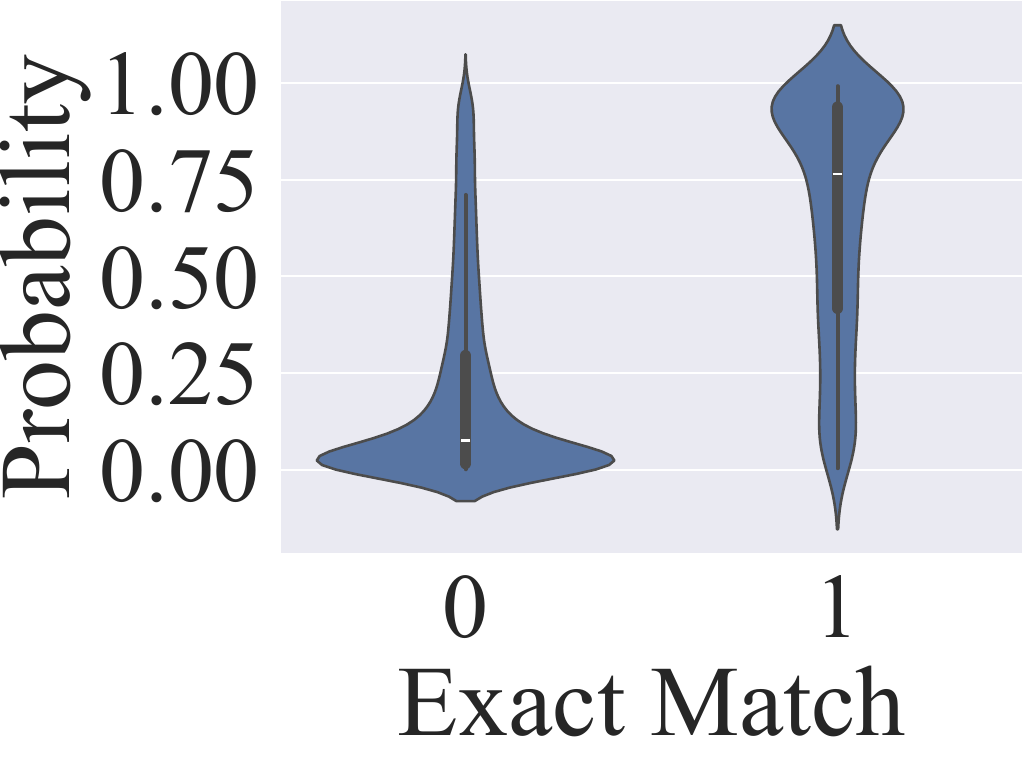} }}\hfill%
    %
    \subfloat[\centering MATH-500 (\texttt{SPARE}-PRM)]{{\includegraphics[width=0.45\columnwidth]{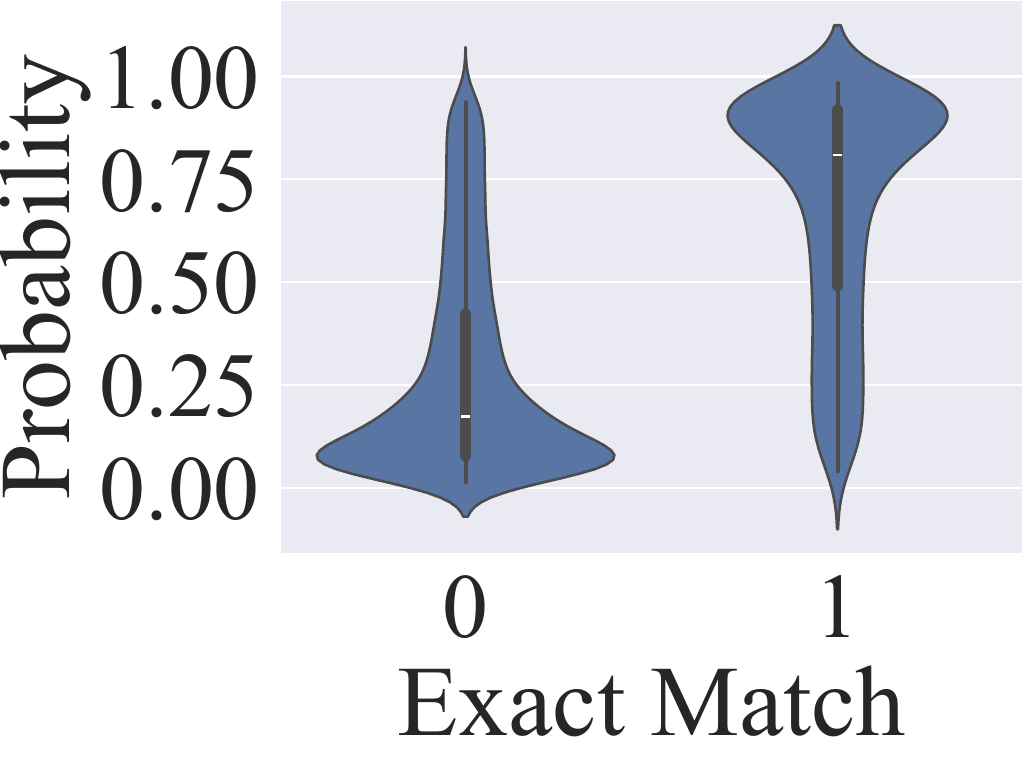} }}\hfill\\
    \subfloat[\centering \texttt{SpaRP-S} (ORM)]{{\includegraphics[width=0.45\columnwidth]{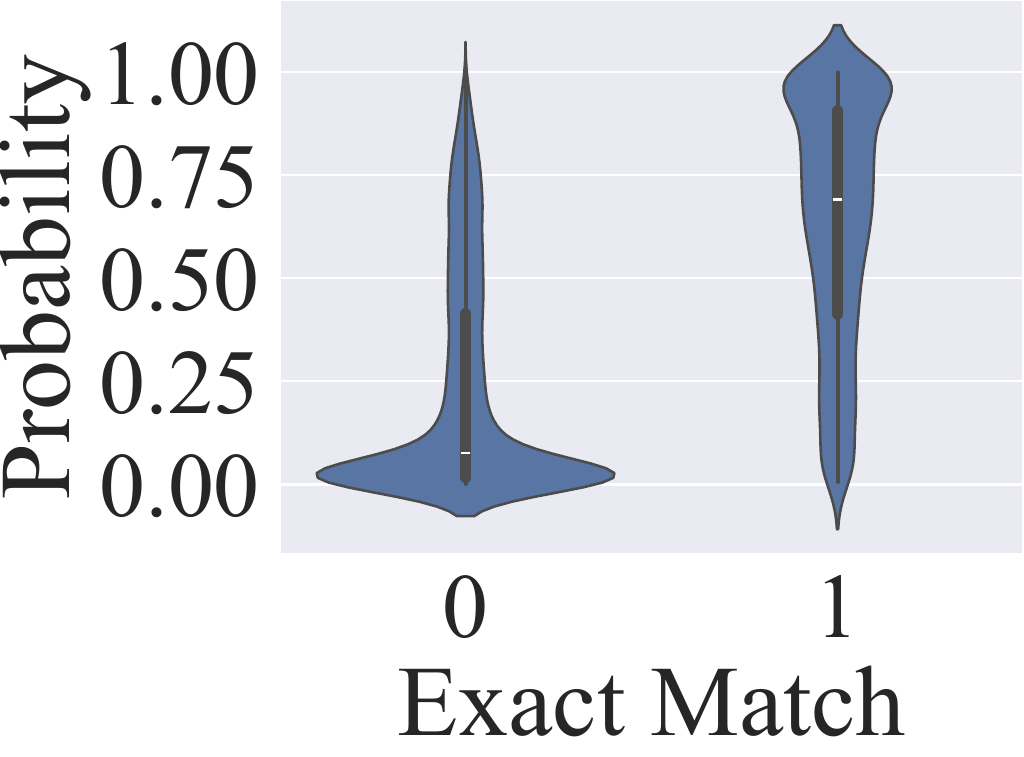} }}\hfill%
    %
    \subfloat[\centering \texttt{SpaRP-S} (\texttt{SPARE}-PRM)]{{\includegraphics[width=0.45\columnwidth]{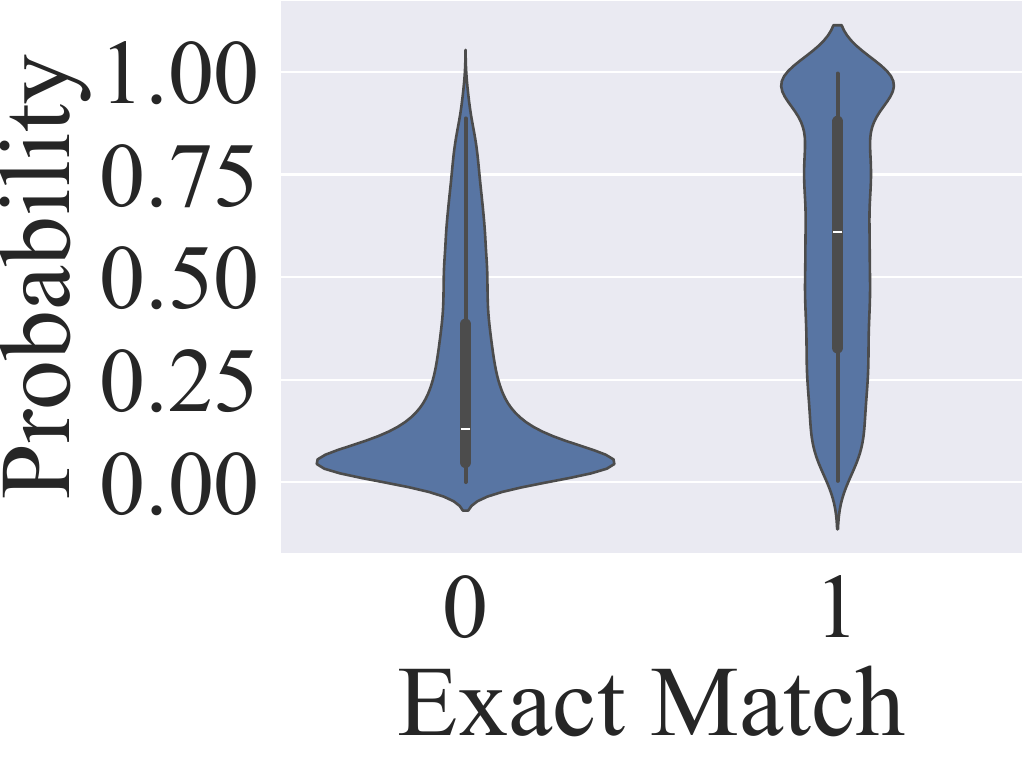} }}\hfill%
    \caption{Distribution-plots of ORM and \texttt{SPARE}-PRM probabilities for correct and incorrect answers of Math-500 and \texttt{SpaRP-S} datasets.}%
    \label{fig:orm-vs-prm-distribution}%
\end{figure}

Table~\ref{tab:contextual_performance} presents additional results on the MATH dataset across varied settings, including different generation counts (N=20 and 256), model families (LLaMA3 and Qwen2.5), and heterogeneous generator–RM configurations (e.g., Qwen2.5-32B generator with Qwen2.5-3B RM), different generation lengths (512 and 2048). \texttt{SPARE} demonstrates consistent performance across these setups, achieving up to a 1.2\% absolute improvement over the ORM baseline. For context, results from Math-Shepherd~\cite{wang-etal-2024-math} and Rest-MCTS$^\ast$~\cite{zhang2024restmcts}, which follow similar methodologies of data construction from scratch and rely on a single model, show comparable gains over ORM.

Finally, Table~\ref{tab:main_results} demonstrates that on datasets with limited reasoning variation (e.g., MuSiQue-Ans, \texttt{SpaRP-S}), \texttt{SPARE}-PRM with Best-of-N (BoN) sampling performs best, while self-consistency (SC) aggregation underperforms even ORMs. In contrast, on datasets with diverse reasoning forms (e.g., GSM8K, MATH-500), SC aggregation boosts \texttt{SPARE}-PRM's performance, in comparison to both BoN and ORM baselines. Distributional analyses (Figure~\ref{fig:orm-vs-prm-distribution}) further confirm this adaptability. On \texttt{SpaRP-S}, \texttt{SPARE}-PRM exhibits wider score spread and lower mean score for correct answers, reducing SC effectiveness. However, on MATH-500, its probability mass skews higher for correct answers, enabling SC to recover stronger performance.

\paragraph{\texttt{SPARE} Helps in Fine tuning.} We report the performance of fine tuning LLM followed by greedy decoding in 
Table~\ref{tab:main_results}. 
\texttt{SPARE}-ORPO achieves the best performance across 
three of the 
four datasets, with a maximum relative improvement of 2.4\% in F1 score on the \texttt{SpaRP-S} dataset compared to the \texttt{Outcome}-ORPO models. On the challenging MATH dataset, it attains a relative improvement of 1.3\%, reaching an accuracy of 23.4\%. The improvements of the \texttt{SPARE} models over \texttt{Outcome} baselines are statistically significant ($p < 0.05$) under one-tailed paired $t$-test. This underscores the effectiveness of \texttt{SPARE} in reasoning step annotation and identifying \textit{superior} preference pairs than outcome-only preference pairs. Both these ORPO models significantly outperform the SFT models trained on the ground-truth reasoning traces, except for the GSM8K dataset. 
We attribute this to the saturation of performance on GSM8K, particularly as we used the same hyperparameters across all datasets (see parameter details in Section~\ref{sec:experiments} and 
Appendix~C. 

\paragraph{\texttt{SPARE} Exhibits Data-Efficient Out-of-Distribution Generalization.} Table~\ref{tab:process_bench} reports a fine-grained evaluation of \texttt{SPARE}-PRMs on the MATH, OlympiadBench, and OmniMATH subsets of the \textsc{ProcessBench}~\cite{zheng-etal-2025-processbench} benchmark, measuring earliest \textit{error} detection, full-solution \textit{correctness}, and their harmonic mean (F1). For comparison, results from the leading PRMs within each model family and size, as well as a PRM trained on human-annotated data (PRM800K), are included from the original benchmark.

\texttt{SPARE}-PRMs achieve the highest accuracy on full-solution correctness identification, significantly outperforming other PRMs on out-of-distribution (OOD) subsets such as OlympiadBench and Omni-MATH. They also consistently match or outperform the PRM trained on human-labeled data. In terms of F1 and earliest error detection, \texttt{SPARE}-Qwen2.5-3B remains comparable on OOD datasets, achieving the highest F1 score on Omni-MATH dataset. Additionally, both Math-Shepherd and \texttt{SPARE}-Qwen2.5-3B exhibit greater robustness to distributional shift, as their error and F1 scores degrade less sharply from MATH to the OOD datasets compared to other top-performing models. Notably, \texttt{SPARE} achieves these results with high data efficiency, using only $\sim$16\% of the training samples compared to both the human-labeled PRM800K and Deepseek-8B synthetic data, and just $\sim$9\% relative to the MCTS-based Math-Shepherd model.

\begin{table}
    \centering
    \begin{tabular}{l c c}
    \toprule
    \textbf{Aggregation} & \textbf{GSM8K} & \textbf{MATH-500} \\
     & Acc. ($\uparrow$) & Acc. ($\uparrow$) \\
    \midrule
    Self-Consistency (SC) & 83.1 & 33.6 \\
    \quad + ORM & 86.7 & 35.1\\
    \quad + Math-Shepherd & \underline{87.7} & \textbf{35.4} \\
    \quad + \texttt{SPARE}-PRM & \textbf{87.8} & \textbf{35.4} \\
    \bottomrule
    \end{tabular}
    \caption{
    Performance comparison on mathematical datasets with Math-Shepherd~\cite{wang-etal-2024-math} MCTS approach. Accuracy averaged over 3 sampling groups.
    }
    \label{tab:mcts_comparison}
\end{table}

\begin{table}
    \centering
    \begin{tabular}{l r r}
    \toprule
    \textbf{Annot. Method} & \textbf{Avg. \#Token} & \textbf{Avg. Time (s)} \\
    \midrule
    MCTS Rollouts (A) & 10,569.1 & 40.5 \\
    \texttt{SPARE} (B) & 4,591.3 & 15.5 \\
    \midrule
    Speed-up (A/B) & 2.3 & 2.6 \\
    Efficiency (100$\times$B/A) & 43.5 & 38.5 \\
    \bottomrule
    \end{tabular}
    \caption{
    \texttt{SPARE} efficiency versus MCTS for process labeling on MATH in terms of average total tokens and runtime under identical compute.
    }
    \label{tab:spare_efficiency}
\end{table}

\paragraph{\texttt{SPARE} is Compute-Efficient and Competitive with MCTS Methods.} For direct comparison with MCTS-based approaches, we follow the experimental setup of Math-Shepherd~\cite{wang-etal-2024-math}, using Mistral-7B:MetaMATH for solution generation (256 samples) and a Mistral-7B-based PRM for \texttt{SPARE}. As shown in Table~\ref{tab:mcts_comparison}, \texttt{SPARE}-PRM slightly outperforms Math-Shepherd on GSM8K and performs comparably on Math-500 under weighted aggregation. Both methods surpass standard baselines such as self-consistency and ORM.

Table~\ref{tab:spare_efficiency} highlights the computational efficiency of \texttt{SPARE} relative to the MCTS-based annotation used in Math-Shepherd. On the MATH dataset, \texttt{SPARE} reduces the average number of total tokens by 2.3$\times$ and runtime on identical compute setup 
(as outlined earlier in Section~\ref{sec:experiments})
by 2.6$\times$, achieving an overall efficiency gain of $\sim$40\% across both metrics. This efficiency arises from \texttt{SPARE}’s single-pass annotation process, in contrast to MCTS-based methods that require extensive search and repeated model inferences, significantly increasing the computational overhead.

\begin{table}
    \centering
    \small
    \begin{tabular}{l l c}
    \toprule
    \textbf{Dataset} & \textbf{Annot. Method} & \textbf{Label Acc.} \\
    \midrule
    \multirow{4}{*}{GSM8K} & DIVERSE-NLI (Llama-based$^\ast$) & 75.6 \\
    & MCTS (Math-Shepherd$^\ast$) & 85.0 \\
    & MCTS (Ours) & 87.3 \\
    & \texttt{SPARE} & 87.5 \\
    \midrule
    \multirow{2}{*}{MATH} & MCTS (Ours) & 76.4 \\
    & \texttt{SPARE} & 76.2 \\
    \bottomrule
    \end{tabular}
    \caption{Step-Label Accuracy of different automatic annotation processes on GSM8K and MATH dataset. Results marked with $^\ast$ are from \citeauthor{wang-etal-2024-math}~\shortcite{wang-etal-2024-math}.}
    \label{tab:label_accuracy}
\end{table}

\paragraph{\texttt{SPARE} Aligns Well with Manual Step Annotations and Exhibits Complementary Behavior to MCTS-Based Annotation.} We assessed the annotation accuracy of \texttt{SPARE} on 56 manually labeled MATH examples, balanced for answer correctness and spanning all seven topics, and 30 examples from GSM8K. Annotations were generated using LLama3-8B in a 5-shot setting, with the target excluded from in-context examples to prevent data leakage. Exemplars were randomly sampled per instance, and each annotation process was repeated ten times. Table~\ref{tab:label_accuracy} reports label accuracies for \texttt{SPARE}, our MCTS implementation, and prior baselines from Math-Shepherd~\cite{wang-etal-2024-math}, where our MCTS follows their setup to ensure fair comparison. On GSM8K, \texttt{SPARE}, MCTS (ours), and MCTS (Math-Shepherd) achieve comparable high accuracy ($\sim$85\%), substantially outperforming DIVERSE-NLI, a directly comparable LLM-based reference-guided approach to \texttt{SPARE}. While DIVERSE-NLI performs alignment externally to single-steps without explicit reasoning, \texttt{SPARE} jointly performs step alignment (including multi-step) and label prediction via explicit reasoning, yielding more reliable annotations, particularly as $\sim$20\% of labeled steps required multi-step alignment. Additionally, despite the increased difficulty of MATH dataset, both \texttt{SPARE} and MCTS maintain strong performance ($\sim$76\%). 

\begin{figure}%
    \centering
    \subfloat[\centering \texttt{SPARE} Single-step]{{\includegraphics[width=0.225\columnwidth]{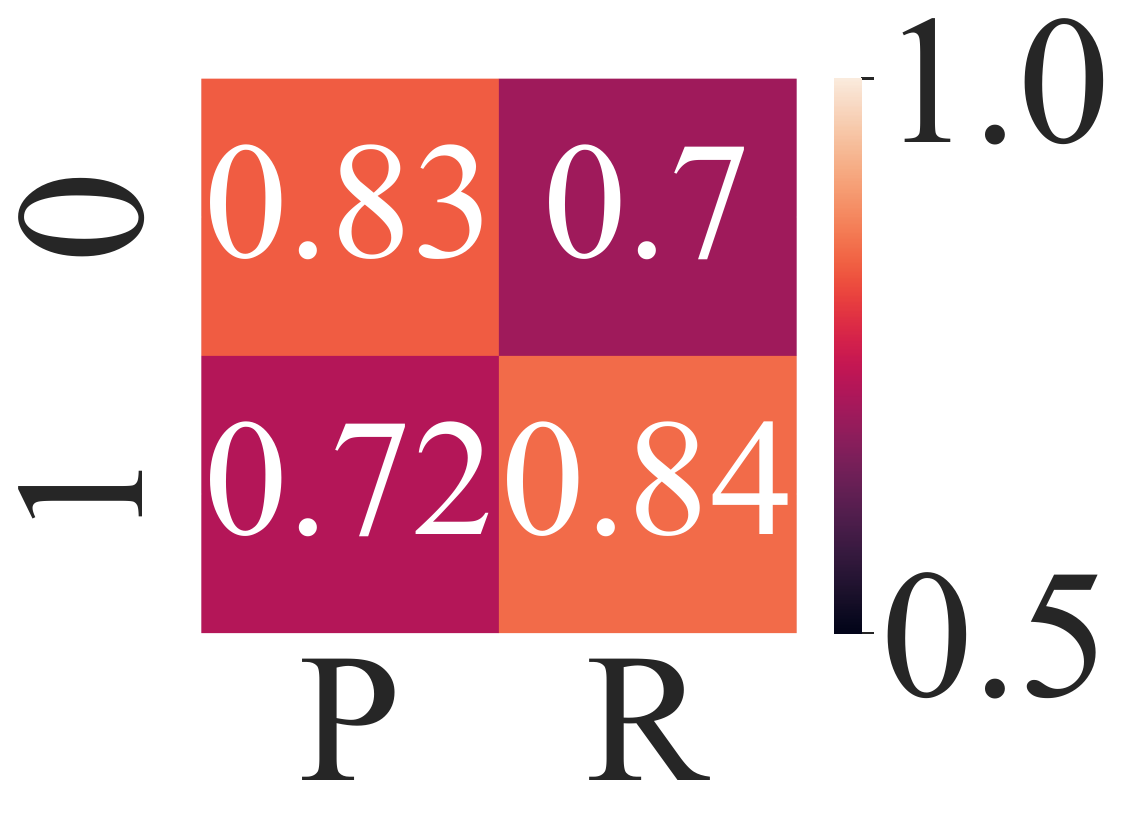} }}\hfill%
    \subfloat[\centering \texttt{SPARE} Multi-step]{{\includegraphics[width=0.225\columnwidth]{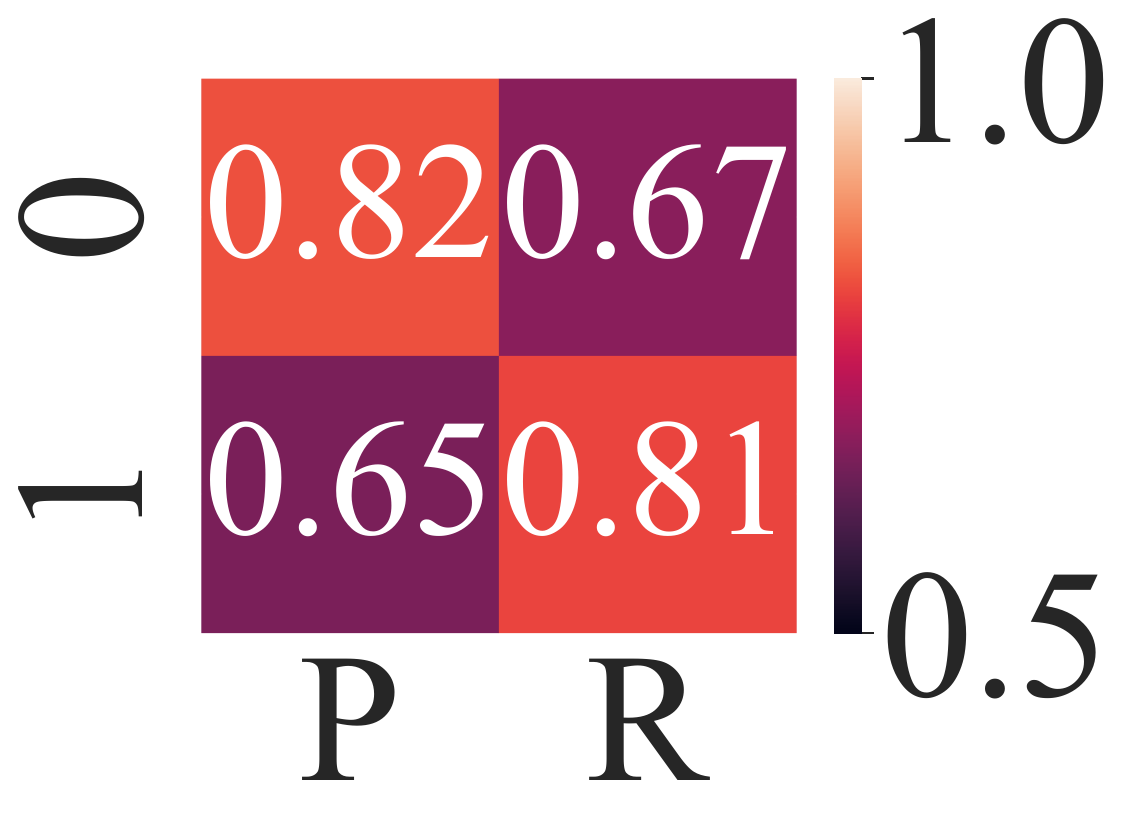} }}\hfill%
    \subfloat[\centering MCTS Single-step]{{\includegraphics[width=0.225\columnwidth]{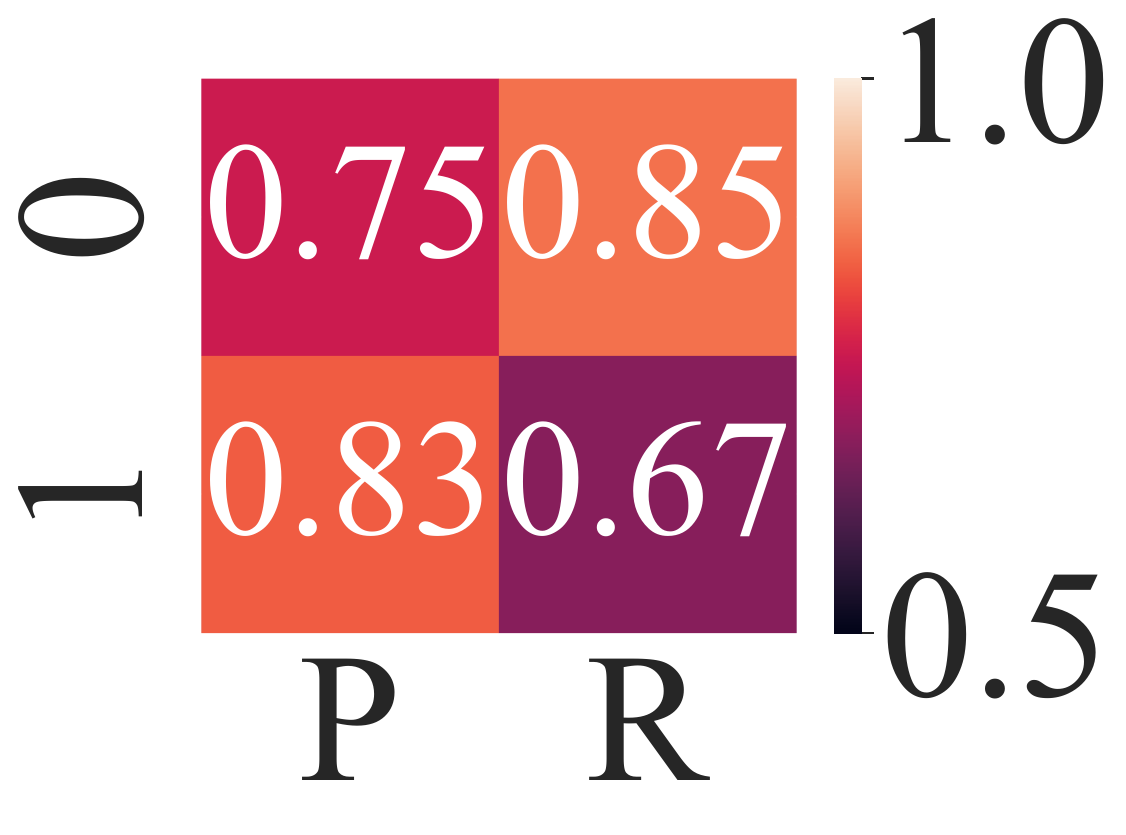} }}\hfill%
    \subfloat[\centering MCTS Multi-step]{{\includegraphics[width=0.225\columnwidth]{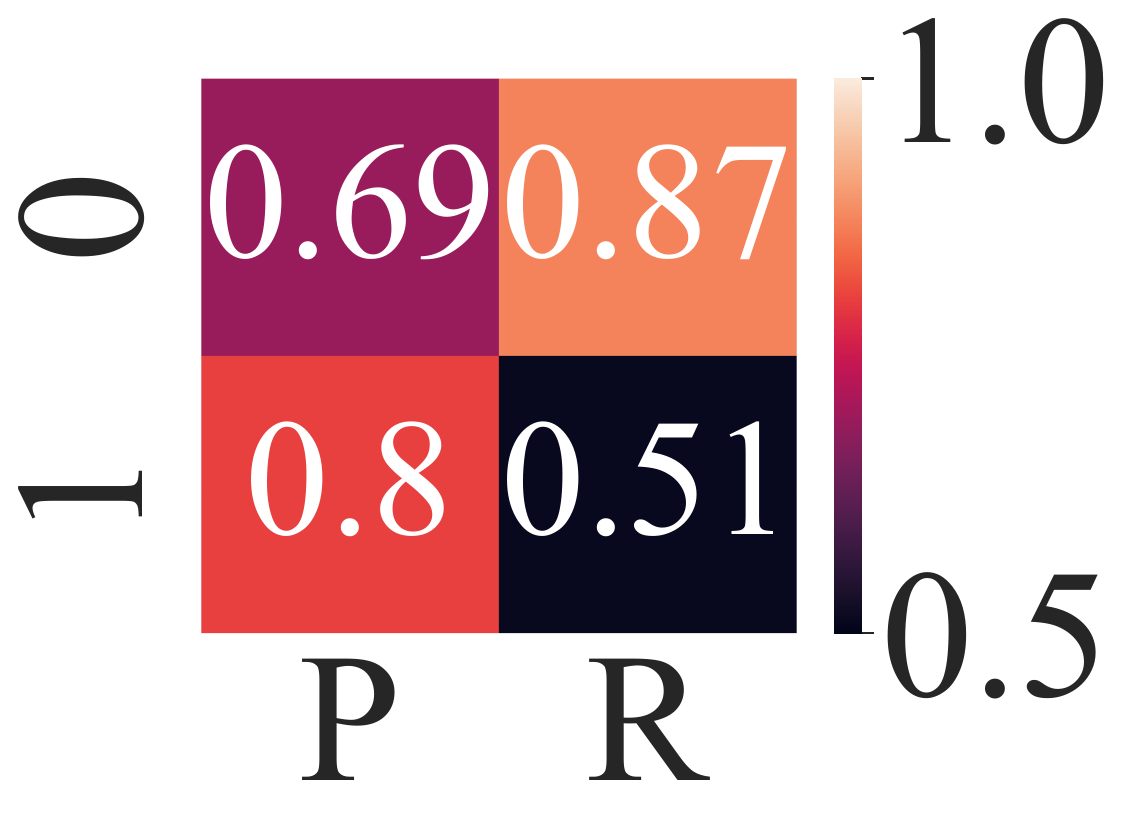} }}%
    \caption{Precision (P) and Recall (R) of \texttt{SPARE} and MCTS annotations relative to human annotations for ground-truth single-step and multi-step alignments.}%
    \label{fig:manual_comparison}%
\end{figure}

Although overall label accuracies for MCTS and \texttt{SPARE} are comparable, the class-wise precision–recall analysis in Figure~\ref{fig:manual_comparison} reveals that for both single- and multi-aligned steps, \texttt{SPARE} achieves high recall for correct steps and high precision for incorrect ones (each $>80\%$), while MCTS excels in precision for correct steps and recall for incorrect ones (each $>80\%$). These complementary strengths suggest potential gains from combining \texttt{SPARE} and MCTS, for example via ensemble annotation. The observed performance drop from single- to multi-aligned steps reflects the increased difficulty of correct assessment of multi-aligned steps.


\section{Conclusion}
\label{sec:conclusion}

We present \textbf{S}ingle-\textbf{P}ass \textbf{A}nnotation with \textbf{R}eference-Guided \textbf{E}valuation (\textbf{\texttt{SPARE}}), a structured framework that enables per-step annotation in a single pass by evaluating each solution step against one or multiple reference steps with explicit reasoning. Our experimental results demonstrate that fine-tuning a base model and training a reward model with \texttt{SPARE} lead to improved reasoning performance under both greedy decoding and ranking/aggregation of multiple solutions. Furthermore, we observe consistent improvements across four datasets spanning mathematical reasoning, multi-hop compositional question answering, and spatial reasoning. \texttt{SPARE} shows (i) data-efficient out-of-distribution generalization on \textsc{ProcessBench}, (ii) competitive performance with greater compute efficiency compared to tree search–based annotation methods, and  (iii) strong alignment with human annotations with complementary precion-recall characteristics. These findings highlight the potential of reference-guided automatic process supervision as a promising approach for enhancing LLM reasoning capabilities.


\section*{Acknowledgments}
This work has been funded by the Collaboration Lab with Nexplore ``AI in Construction'' (AICO). We further thank our anonymous reviewers and Chen Cecilia Liu, Rima Hazra, and Tim Baumgärtner for their fruitful discussions and helpful feedback.

\bibliography{aaai2026, anthology_p1, custom}


\appendix

\section{\texttt{SPARE} Annotation Example}
\label{sec:grading_example}

\begin{figure*}
    \centering
    \includegraphics[width=\linewidth]{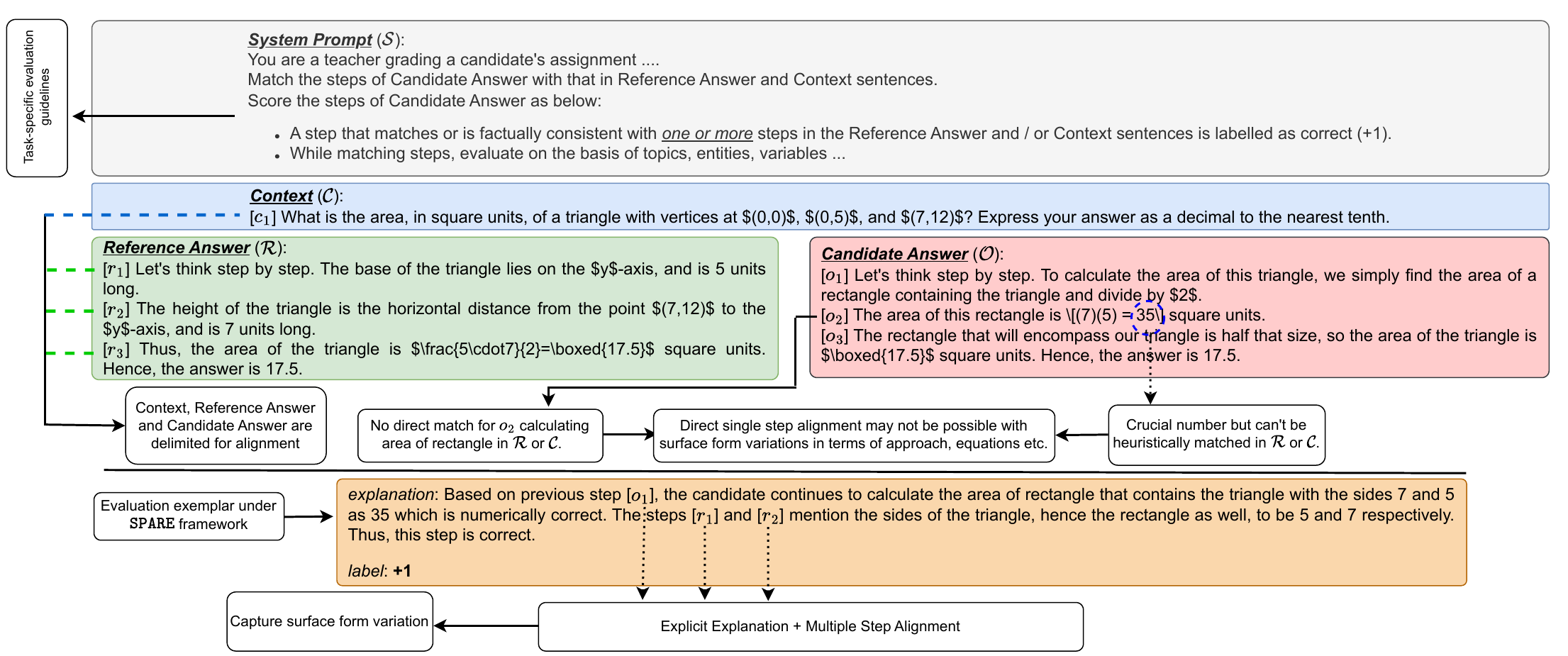}
    \caption{Determining the correctness of a step in a candidate solution against a given context and reference solution presents several challenges, including step alignment, surface-form variations, and heuristic limitations. We propose a unified, single-stage framework: \texttt{\textbf{S}}ingle \texttt{\textbf{P}}ass \texttt{\textbf{A}}nnotation with \texttt{\textbf{R}}eference-Guided \texttt{\textbf{E}}valuation (\textbf{\texttt{SPARE}}: $(\mathcal{S, C, R, O}) \rightarrow \mathcal{E}$). \texttt{SPARE} produces an explanation-based step-by-step evaluation $\mathcal{E}$ of a candidate model output $\mathcal{O}$, grounded to a given context $\mathcal{C}$, reference reasoning $\mathcal{R}$, and system prompt $\mathcal{S}$ (Section~\ref{sec:SPARE} \& Figure~\ref{fig:SPARE_overview}).}
    \label{fig:spare-example-overview}
\end{figure*}

\begin{figure}
\centering

\subfloat[\textit{One-to-one}]{%
\centering
\begin{tikzpicture}[
    baseline=(current bounding box.center),
    every node/.style={circle, draw, minimum size=0.5cm},
    >=stealth
]
\node (rj) at (0,0) {$r_j$};
\node (oi) at (2.5,0) {$o_i$};
\coordinate (merge) at (1.95,0);

\draw (rj) -- (merge);
\draw[->] (merge) -- (oi);
\end{tikzpicture}
}
\quad
\subfloat[\textit{One-to-many}]{%
\centering
\begin{tikzpicture}[
    baseline=(current bounding box.center),
    every node/.style={circle, draw, minimum size=0.5cm},
    >=stealth
]
\node (ck) at (0,1.25) {$c_k$};
\node (rj) at (0,0) {$r_j$};
\node (ol) at (0,-1.25) {$o_l$};
\node (oi) at (2.5,0) {$o_i$};
\coordinate (merge) at (1.5,0);

\draw (ck) -- (merge);
\draw (rj) -- (merge);
\draw (ol) -- (merge);
\draw[->] (merge) -- (oi);
\end{tikzpicture}
}

\caption{Conceptual illustration of different alignment scenarios.}
\label{fig:alignment_scenario}
\end{figure}

An example overview of the \texttt{SPARE} framework is presented in Figure~\ref{fig:spare-example-overview}. Furthermore, an example system prompt with evaluation guidelines and user request for MATH dataset under our \texttt{SPARE} framework is provided in Table~\ref{tab:grading_system_user}. A sample graded output and the list of error categories is provided in Table~\ref{tab:grading_assistant} and Table~\ref{tab:error_categories}, respectively. Finally, as LLM-based grading is inherently noisy, \texttt{SPARE} exhibits some failure modes that yield incorrect annotations, examples of which are presented in Table~\ref{tab:failure_modes}. Other datasets have their own domain and subject specific evaluation heuristics included in the system prompt.

A conceptual illustration of \textit{one-to-one} and \textit{one-to-many} alignment scenario from Section~\ref{sec:SPARE} is presented in Figure~\ref{fig:alignment_scenario}. A discussion with representative examples is further presented below:

\paragraph{One-to-one alignment.} 
Direct and complete correspondence between two steps, say, a reference step and a model-generated step. The two steps express the same underlying operation or conclusion but may differ in surface form or phrasing. For example, a reference step, say, 
\emph{R2: Compute $3 \times 4 = 12$} aligns exactly with a model output step, say, 
\emph{O4: 3 times 4 equals 12}. Although the wording differs slightly, both steps perform the identical computation and reach the same intermediate result, constituting a one-to-one alignment.

\paragraph{One-to-many alignment.} 
A single composite step corresponds to at least two steps, such that a composite reasoning or computation in one step is spread across two or more steps in the other. For example, the reference steps:

\begin{itemize}
    \item \emph{R2: 10 -- 4 = 6 candies remain with Jake after giving 4 candies to his friend.}
    \item \emph{R3: 6 + 6 = 12 candies with Jake after he buys 6 more candies.}
\end{itemize}

together correspond to the model output step:

\begin{itemize}
    \item \emph{O1: Accounting for the candies that he bought and the ones that he gave away, Jake now has $10 + 6 - 4 = 12$ candies.}
\end{itemize}

Although the reference breaks the computation into two steps, the model compresses the reasoning into a single, composite step, illustrating a one-to-many alignment for a correct reference-guided evaluation.

\newcommand{\minipagewidth}{\textwidth}
\newcommand{\termwidth}{0.1\textwidth}
\begin{table*}[]
    \centering
    \small
    \begin{adjustbox}{max width=0.9\textwidth}
    \begin{tabular}{c c}
        \toprule
        \textbf{Role} & \textbf{Content} \\
        \midrule
        \begin{minipage}{\termwidth} System \end{minipage} & \begin{minipage}{\minipagewidth} You are a teacher grading a student's assignment. You are given a QUESTION, its ground-truth correct REFERENCE ANSWER and a STUDENT'S ANSWER. You are asked to match the steps of STUDENT'S ANSWER with that in the REFERENCE ANSWER and in the context of the given QUESTION. You are required to score the steps of STUDENT'S ANSWER as below:
        \\ \\
        A step in the STUDENT'S ANSWER that matches or is factually consistent with one or more steps in the REFERENCE ANSWER and in the context of the sentences provided in the QUESTION is labelled as CORRECT. While matching steps, evaluate on the basis of:
        \\ \\
        (a) whether the topic, entities, variables and the intended result of the step are correct or not, and
        \\
        (b) whether the expressions, equations, and / or numerical computations in a step are correct or not.
        \\ \\
        A step in the STUDENT'S ANSWER that doesn't match or is factually incorrect with respect to the provided REFERENCE ANSWER and the QUESTION is labelled as INCORRECT.
        \\ \\
        You need to evaluate ALL the steps of the STUDENT'S ANSWER. Provide your evaluation ONLY and ONLY in JSON format as a list of dictionaries whose keys and their intended purposes are:
        \\ \\
        ``student\_step'': The current step number of the STUDENT'S ANSWER.
        \\ \\
        ``reasoning'': The reasoning expanding upon why or what part of the current `student\_step` of the STUDENT'S ANSWER, either DIRECTLY and ENTIRELY in itself or probably in combination with other steps in the STUDENT'S ANSWER, is correct or incorrect in reference to one or more REFERENCE ANSWER steps and the QUESTION sentences.
        \\ \\
        ``question\_sentences'': A list of sentences in the QUESTION based on which the correctness or the incorrectness of the current `student\_step` is reasoned and arrived at. If the number of steps in the STUDENT'S ANSWER is less than that in the REFERENCE ANSWER or the topic and the intended goal of a `student\_step` doesn't match with any steps in the REFERENCE ANSWER, then it is useful to look for QUESTION sentences to assess the `student\_step`. More than one QUESTION sentences can be of use for this evaluation.  Leave it as an empty list if the current `student\_step` DIRECTLY and ENTIRELY matches with one or more steps in the REFERENCE ANSWER.
        \\ \\
        ``student\_combining\_steps'': A list of previous `student\_step` that is necessary in evaluating the current `student\_step` because of restatements or transformations, or when combined with the current `student\_step` will be part or whole of one or more steps in the REFERENCE ANSWER. Leave it as an empty list if the current `student\_step` DIRECTLY and ENTIRELY matches with one or more steps in the REFERENCE ANSWER. If the number of steps in the STUDENT'S ANSWER is more than that in the REFERENCE ANSWER, then a single step in REFERENCE ANSWER may correspond to multiple steps in the STUDENT'S ANSWER and this list will be non-empty for some of the `student\_step`.
        \\ \\
        ``matching\_reference\_steps'': A list of steps in the REFERENCE ANSWER based on which the correctness or the incorrectness of the current `student\_step` is reasoned and arrived at. If the number of steps in the STUDENT'S ANSWER is less than that in the REFERENCE ANSWER, then multiple steps in the REFERENCE ANSWER may correspond to a single step in the STUDENT'S ANSWER. Leave this empty if there are no matching steps in the REFERENCE ANSWER.
        \\ \\
        ``error\_category'': A list of type of errors that caused the current `student\_step` to be partially or fully incorrect. Report from ``COMPREHENSION'' when the student misunderstands and misapplies some concept, ``NUMERIC'' when the numeric values don't match, ``CALCULATION'' when the computations done are incorrect, ``TRANSFORMATION'' when the algebraic or trigonometric rearrangements or substitutions are incorrect, ``PROPAGATION'' when the incorrectness gets carried forward directly from the previous `student\_step`, ``RESTATEMENT'' when the student makes mistake in restating from its own or question sentences, and ``NO STEP MATCH''. Leave it as an empty list if the current `student\_step` is completely correct.
        \\ \\
        ``label'': binary score of the current `student\_step` as either CORRECT or INCORRECT. \end{minipage}\\
        \midrule
        \begin{minipage}{\termwidth} User \end{minipage} & \begin{minipage}{\minipagewidth} QUESTION:
        \\
        \textnormal{[1]} My three friends and I have dinner together every weekend.
        \\
        \textnormal{[2]} Each weekend, two of us cook and the other two clean up afterwards.
        \\
        \textnormal{[3]} How many different ways are there for us to choose who cooks and who cleans?
        \\ \\
        REFERENCE ANSWER:
        \\
        \textnormal{[1]} Let's think step by step.\textbackslash nThere are four ways to choose the first cook and three ways to choose the second cook, but this counts every pair of cooks twice since order doesn't matter.
        \\
        \textnormal{[2]} Once the cooks are chosen, the two people left over are the cleaners.
        \\
        \textnormal{[3]} So, there are $(4\cdot 3)/2=\boxed{6}$ ways for us to choose who cooks and who cleans.\textbackslash nHence, the answer is 6.
        \\ \\
        STUDENT'S ANSWER:
        \\
        \textnormal{[1]} Let's think step by step.\textbackslash nWe have four choices for who cooks, and then we have three choices for who cleans.
        \\
        \textnormal{[2]} Since each order cooks, it doesn't matter who cleans; only the total cleaning pair counts.
        \\
        \textnormal{[3]} There are $\dbinom{4}{2}\cdot\dbinom{2}{1}=\boxed{12}$ ways in total.\textbackslash nHence, the answer is 12.
        \end{minipage} \\
        \bottomrule
    \end{tabular}
    \end{adjustbox}
    \caption{An example system prompt with evaluation guidelines and evaluation task as user prompt for MATH dataset under our \texttt{SPARE} framework. Other datasets have their own domain and subject specific evaluation details included in the system prompt.}
    \label{tab:grading_system_user}
\end{table*}

\renewcommand{\minipagewidth}{\textwidth}
\renewcommand{\termwidth}{0.1\textwidth}
\begin{table*}[]
    \centering
    \small
    \begin{adjustbox}{max width=0.9\textwidth}
    \begin{tabular}{c c}
        \toprule
        \textbf{Role} & \textbf{Content} \\
        \midrule
        \begin{minipage}{\termwidth} Assistant \end{minipage} & \begin{minipage}{\minipagewidth}\verb|```|json \\
        \textnormal{[} \\
        \hspace*{1em} \{ \\
        \hspace*{2em} ``student\_step'': 1, \\
        \hspace*{2em} ``reasoning'': The step [1] of the STUDENT'S ANSWER mentions that there are 3 choices for who cleans. Sentence [2] of \\ \hspace*{4em}QUESTION and step [2] of REFERENCE ANSWER mention that there two people who can clean. Hence this is a \\ \hspace*{4em}mismatch and the student's assertion is wrong., \\
        \hspace*{2em} ``question\_sentences'': [2], \\
        \hspace*{2em} ``student\_combining\_steps'': [], \\
        \hspace*{2em} ``matching\_reference\_steps'': [2], \\
        \hspace*{2em} ``error\_category'': [``COMPREHENSION''], \\
        \hspace*{2em} ``label'': [``INCORRECT''], \\
        \hspace*{1em} \} \\
        \hspace*{1em} \{ \\
        \hspace*{2em} ``student\_step'': 2, \\
        \hspace*{2em} ``reasoning'': The step [2] of the STUDENT'S ANSWER is a generic restatement of the fact that the count of cleaners \\ \hspace*{4em} follows automatically from the count of cooks., \\
        \hspace*{2em} ``question\_sentences'': [], \\
        \hspace*{2em} ``student\_combining\_steps'': [], \\
        \hspace*{2em} ``matching\_reference\_steps'': [], \\
        \hspace*{2em} ``error\_category'': [], \\
        \hspace*{2em} ``label'': [``CORRECT''], \\
        \hspace*{1em} \} \\
        \hspace*{1em} \{ \\
        \hspace*{2em} ``student\_step'': 3, \\
        \hspace*{2em} ``reasoning'': The step calculates the total ways or count to be 12 which is twice the ground truth ways of 6 as reported in \\ \hspace*{4em} [3] of the REFERENCE ANSWER. This could be because of the overuse of the factor 2 in doing the calculations due \\ \hspace*{4em} to the misunderstandings in the previous steps., \\
        \hspace*{2em} ``question\_sentences'': [], \\
        \hspace*{2em} ``student\_combining\_steps'': [], \\
        \hspace*{2em} ``matching\_reference\_steps'': [3], \\
        \hspace*{2em} ``error\_category'': [``COMPREHENSION'', ``CALCULATION''], \\
        \hspace*{2em} ``label'': [``INCORRECT''], \\
        \hspace*{1em} \} \\
        \textnormal{]} \\
        \verb|```| \end{minipage} \\
        \bottomrule
    \end{tabular}
    \end{adjustbox}
    \caption{A graded example for MATH dataset under our \texttt{SPARE} framework.}
    \label{tab:grading_assistant}
\end{table*}

\renewcommand{\minipagewidth}{\textwidth}
\renewcommand{\termwidth}{0.1\textwidth}
\begin{table*}
    \centering
    \small
    \begin{adjustbox}{max width=0.9\textwidth}
    \begin{tabular}{c c}
        \toprule
        \textbf{Dataset} & \textbf{Error Categories} \\
        \midrule
        \begin{minipage}{\termwidth} MATH \end{minipage} & \begin{minipage}{\minipagewidth} \begin{enumerate}
            \item \textbf{Comprehension} -- The student misunderstands and misapplies some concept.
            \item \textbf{Numeric} -- The numeric values don't match.
            \item \textbf{Calculation} -- The computations done are incorrect. \item \textbf{Transformation} -- The algebraic or trigonometric rearrangements or substitutions are incorrect.
            \item \textbf{Propagation} -- The incorrectness gets carried forward directly from previous student steps. 
            \item \textbf{Restatement} -- The student makes mistake in restating from its own or question sentences.
            \item \textbf{No Step Match} -- The step doesn't match with any reference step nor it can be inferred from any context / question sentences.
        \end{enumerate}
        \end{minipage} \\
        \midrule
        \begin{minipage}{\termwidth} GSM8K \end{minipage} & \begin{minipage}{\minipagewidth} \begin{enumerate}
            \item \textbf{Comprehension} -- The student misunderstands and misapplies some concept.
            \item \textbf{Numeric} -- The numeric values don't match.
            \item \textbf{Calculation} -- The computations done are incorrect. \item \textbf{No Step Match} -- The step doesn't match with any reference step nor it can be inferred from any context / question sentences.
        \end{enumerate}
        \end{minipage} \\
        \midrule
        \begin{minipage}{\termwidth} 
        MuSiQue-Ans \end{minipage} & \begin{minipage}{\minipagewidth} \begin{enumerate}
            \item \textbf{Document Name} -- The document name in the step doesn't match or exists.
            \item \textbf{Entity Name} -- The entity name in the step doesn't match or exists.
            \item \textbf{Numeric} -- The numbers mentioned don't match.
            \item \textbf{Intended Category} -- The category differs from what is required e.g. Date required but Country discussed.
            \item \textbf{Semantic Relation} -- The relation between entities, to be compared semantically e.g. local language and native language in reference to a person and a place is semantically same.
            \item \textbf{No Step Match} -- The step doesn't match with any reference step nor it can be inferred from any context / question sentences.
        \end{enumerate}
        \end{minipage} \\
        \midrule
        \begin{minipage}{\termwidth} \texttt{SpaRP} \end{minipage} & \begin{minipage}{\minipagewidth} \begin{enumerate}
            \item \textbf{Entity Name} -- The name of the entity in the step doesn't match or exist.
            \item \textbf{Incorrect Relation} -- The reported spatial relation in the step doesn't match or exist.
            \item \textbf{No Step Match} -- The step doesn't match with any reference step nor it can be inferred from any context / question sentences.
        \end{enumerate}
        \end{minipage} \\
        \bottomrule
    \end{tabular}
    \end{adjustbox}
    \caption{Dataset specific error categories.}
    \label{tab:error_categories}
\end{table*}

\renewcommand{\minipagewidth}{\textwidth}
\begin{table*}[]
    \centering
    \small
    \begin{adjustbox}{max width=0.9\textwidth}
    \begin{tabular}{l}
        \toprule
        \multicolumn{1}{c}{\textbf{Failure Modes and Examples}} \\
        \midrule
        \begin{minipage}{\minipagewidth}\textbf{Failure Mode}: Disregards reference and perform incorrect evaluation independently.\\  \textbf{Reference Solution}: \textnormal{[1]} Let's think step by step. The decimal representation of $\frac{6}{7}$ is $0.\overline{857142}$, which repeats every 6 digits.\textbackslash n\textnormal{[2]} Since 100 divided by 6 has a remainder of 4, the 100th digit is the same as the fourth digit following the decimal point, which is $\boxed{1}$.\textbackslash n Hence, the answer is 1.\\ \textbf{Generated Solution}: \textnormal{[1]} Let's think step by step.\textbackslash n Note that all but a finite number of digits in the decimal expansion of any fraction that has a non-repeating decimal expansion are repeated in a pattern.\textbackslash n \textnormal{[2]} In the decimal expansion of $\frac{6}{7}$, we have the repeating pattern of 857142, which has a length of 6.\textbackslash n \textnormal{[3]} This implies that the 100th and the 106th, 112th, 118th, and all other digits that are 6 or 2 more than a multiple of 6 all have the same value.\textbackslash n \textnormal{[4]} The third digit after the decimal that is a multiple of 6 is the 6th, and we know that 6 is 0 from the pattern.\textbackslash n \textnormal{[5]} Therefore, the 100th is $\boxed{0}$ as well.\textbackslash n Hence, the answer is 0.\\ \textbf{Manual Grading}: \{``student\_step'': 4, ``reasoning'': ``The step [4] of the STUDENT'S ANSWER works out that the third digit after the decimal in the repeating set of digits is required for answering. The step [2] of the REFERENCE ANSWER mentions that fourth digit after the decimal will lead to the answer. Hence, there is a mismatch.'', ``question\_sentences'': [], ``student\_combining\_steps'': [], ``matching\_reference\_steps'': [2], ``error\_category'': [``NUMERIC''], ``label'': ``INCORRECT''\} \\ \textbf{LLM Grading}: \{``student\_step'': 4, ``reasoning'': ``The step [4] of the STUDENT'S ANSWER correctly identifies the 6th digit after the decimal point as 0.'', ``question\_sentences'': [], ``student\_combining\_steps'': [], ``matching\_reference\_steps'': [1], ``error\_category'': [], ``label'': ``CORRECT''\}\end{minipage} \\
        \midrule
        \begin{minipage}{\minipagewidth}\textbf{Failure Mode}: Incorrect step count identification resulting in partial or excessive evaluation.\\ \textbf{Generated Solution}: \textnormal{[1]} Let's think step by step.\textbackslash n First, note that the three sides that the square of side $y$ units has will have a length of $y$ units.\textbackslash n [2] Second, note that the four congruent rectangles are congruent, so it does not matter which rectangle we consider.\textbackslash n [3] Therefore, the area of one of the congruent rectangles is $xy-y^2=x(y-x)$.\textbackslash n [4] The sides of the rectangle have lengths $x$ and $y-x$, so its perimeter is $2x+2(y-x)=2x+2y-2x=2y$.\textbackslash n [5] Therefore, the perimeter of the congruent rectangle is $\boxed{2y}$ units.\textbackslash n Hence, the answer is 2y. \\ \textbf{LLM Grading}: [\{``student\_step'': 1, ``reasoning'': ``The step [1] of the STUDENT'S ANSWER talks about the sides of the square of side $y$ units. ...''\}, ...  \{``student\_step'': 3, ``reasoning'': ``The step [3] of the STUDENT'S ANSWER calculates the area of one of the congruent rectangles as $xy-y^2=x(y-x)$. ...''\}] \# Evaluation terminates prematurely at step 3 despite solution extending to step 5.
        \end{minipage} \\
        \midrule
        \begin{minipage}{\minipagewidth}\textbf{Failure Mode}: Conceptual difference in evaluation granularity or strictness criteria. \\ \textbf{Reference Solution}: [1] Let's think step by step.\textbackslash n There are four ways to choose the first cook and three ways to choose the second cook, but this counts every pair of cooks twice since order doesn't matter.\textbackslash n [2] Once the cooks are chosen, the two people left over are the cleaners.\textbackslash n [3] So, there are $(4\cdot 3)/2=\boxed{6}$ ways for us to choose who cooks and who cleans.\textbackslash n Hence, the answer is 6. \\ \textbf{Generated Solution}: [1] Let's think step by step.\textbackslash n We have four choices for who cooks, and then we have three choices for who cleans.\textbackslash n [2] Since each order cooks, it doesn't matter who cleans; only the total cleaning pair counts.\textbackslash n [3] There are $\dbinom{4}{2}\cdot\dbinom{2}{1}=\boxed{12}$ ways in total.\textbackslash n Hence, the answer is 12. \\ \textbf{LLM Grading}: \{``student\_step'': 1, ``reasoning'': ``The step [1] of the STUDENT'S ANSWER states that there are four choices for who cooks. The step [1] of the REFERENCE ANSWER also mentions that there are four ways to choose the first cook. Hence, this step is a complete match.'', ``question\_sentences'': [], ``student\_combining\_steps'': [], ``matching\_reference\_steps'': [1], ``error\_category'': [], ``label'': ``CORRECT''\} \# Applied lenient evaluation by overlooking the emphasis on \textit{first} cook among four choices in the reference solution.
        \end{minipage} \\
        \bottomrule
    \end{tabular}
    \end{adjustbox}
    \caption{Examples of \texttt{SPARE} Annotation Failures Modes. For brevity, only relevant manual and LLM-graded step annotations are presented.}
    \label{tab:failure_modes}
\end{table*}

\section{Details of Reward Model (RM) Training}
\label{sec:reward_model_details}

The number $N$ of positive and negative samples (i.e. $N/2$ pairs) for Reward Model training are presented in Table~\ref{tab:rm_training_detail}. Hence, for example, the total effective outcome or process supervision dataset used for RM training for mathemtical datasets was $\approx$40K. This is still significantly smaller than other work~\cite{lightman2024lets, wang-etal-2024-math}.

\begin{table}[]
    \centering
    \begin{tabular}{c c}
        \toprule
        Dataset & N \\
        \midrule
        GSM8K & 40,350 \\ 
        MATH & 40,500 \\ 
        MuSiQue & 10,000 \\ 
        SpaRP & 16,000 \\ 
        \bottomrule
    \end{tabular}
    \caption{Training data sizes for Reward Models}
    \label{tab:rm_training_detail}
\end{table}

While prior work~\cite{lightman2024lets, wang-etal-2024-math, luo2024improvemathematicalreasoninglanguage} have shown the $\operatorname{min}$ or $\operatorname{prod}$ aggregation to be the better performing aggregation strategies, other work~\cite{wang-etal-2024-multi-step} have reported these to underperform ORM when the annotation process differs. For their annotation process, they reported $\operatorname{last}$ aggregation strategy, among others, to outperform ORM. We also found the $\operatorname{min}$ and $\operatorname{prod}$ aggregation strategies to sometimes underperform the ORMs, while $\operatorname{last}$ aggregation strategy performing the best. Hence, all the metrics are reported using the $\operatorname{last}$ aggregation strategy for PRMs. 

\begin{table}
    \centering
    \small
    \begin{adjustbox}{max width=\columnwidth}
    \begin{tabular}{c c}
        \toprule
        \textbf{Training} & \textbf{\texttt{GSM8K}} \\
        \textbf{Method} & Acc. $(\uparrow)$ \\
        \midrule
        SFT $1^{st}$ Iteration & 68.61 \\
        \quad + $2^{nd}$ Iteration & 70.36 \\
        \quad + \texttt{Out.}-ORPO & \underline{71.27} \\
        \midrule
        \quad + \texttt{SPARE}-ORPO & \textbf{72.25} \\
        \bottomrule
    \end{tabular}
    \end{adjustbox}
    \caption{Performance evaluations of Llama-3 8B Instruct model with \textit{greedy decoding} under different training methods on the GSM8K dataset with $lr=10^{-5}$ for $1^{st}$ SFT. followed by $lr=5\times10^{-6}$ for other iterations. Best values in \textbf{bold}, second best in \underline{underline}.}
    \label{tab:gsm8k_finetuning}
\end{table}

\section{Details of Finetuning}
\label{sec:finetuning_details}

\subsection{Implementation detail common across all datasets.}
\label{sec:common_hyperparameters}
We used Huggingface's TRL library and QLoRA for parameter-efficient finetuning of all ORPO models across datasets, 
using a fixed set of hyperparameters for consistency (Table~\ref{tab:finetuning_parameters}). This setup performed well across all datasets except GSM8K, where we observed performance saturation after the first iteration and degradation in ORPO models. To address this, we conducted a learning rate search specifically for GSM8K, leading to consistent improvements as detailed in Section~\ref{sec:gsm8k_hyperparameters}.

\subsection{GSM8K Hyperparameter Tuning.}
\label{sec:gsm8k_hyperparameters}

Due to training saturation observed on GSM8K with the common hyperparameters (Table~\ref{tab:finetuning_parameters}), we conducted a targeted search over learning rates $\{10^{-6}, 5\times10^{-6}, 10^{-5}, 5\times10^{-5}, 10^{-4}\}$. Using $lr = 10^{-5}$ for the first iteration and $lr = 5 \times 10^{-6}$ for the second on \texttt{SPARE} produced the best results on the validation/development dataset. The corresponding test set performance is reported in Table~\ref{tab:gsm8k_finetuning}, where the \texttt{SPARE}-ORPO model achieved an accuracy of 72.25\%. We did not include it in 
the main results table 
in order to ensure a robust comparison by using the same hyperparameters across all datasets.  

\begin{table}[]
    \centering
    \begin{tabular}{c c}
        \toprule
        \textbf{Parameter Name} & \textbf{Value} \\
        \midrule
        QLoRA: &  \\
        \midrule
        $\alpha$ & 16 \\
        Dropout & 0.1 \\
        $r$ & 64 \\
        bias & None \\
        task\_type & CAUSAL\_LM \\
        \midrule
        Training Arguments: & \\
        \midrule
        Effective Batch Size & 32 \\
        $lr$ & $1.0e-4$ \\
        weight decay & 0.001 \\
        max\_grad\_norm & 0.3 \\
        warm up ratio & 0.03 \\
        lr\_scheduler & cosine \\
        \bottomrule
    \end{tabular}
    \caption{Values of the parameters and hyperparameters used while ORPO finetuning.}
    \label{tab:finetuning_parameters}
\end{table}

\section{Pointwise vs Pairwise ORMs.} 
\label{sec:point_vs_pair}

\begin{table}
    \centering
    \small
    \begin{adjustbox}{max width=\columnwidth}
    \begin{tabular}{l c c c c}
        \toprule
         & \multicolumn{2}{c}{\textbf{Mathematical}} & \textbf{Question} & \textbf{Spatial} \\
          & \multicolumn{2}{c}{\textbf{Reasoning}} & \textbf{Reasoning} & \textbf{Reasoning} \\
         \cmidrule{2-5}
        \textbf{Aggregation /} & \textbf{\texttt{GSM8K}} & \textbf{\texttt{MATH-500}} & \textbf{
        \texttt{MuSiQue-Ans}} & \textbf{\texttt{SpaRP-S}} \\
        \textbf{Ranking} & Acc. $(\uparrow)$ & Acc. $(\uparrow)$ & Acc. $(\uparrow)$ / F1 $(\uparrow)$ & Acc. $(\uparrow)$ / F1 $(\uparrow)$ \\
        \midrule
        \textit{pair.}-ORM & 78.54 & 16.30 & 30.45 / 42.87 & 31.65 / 40.78 \\
        \textit{pair.}-ORM + SC & 79.45 & 21.00 & 34.13 / 43.82 & 31.63 / 40.77 \\
        \texttt{SPARE}-ORM & 79.15 & 19.00 & 34.67 / 45.11 & 32.7 / 41.95 \\
        \texttt{SPARE}-ORM + SC & 79.22 & 20.60 & 35.29 / 45.52 & 32.63 / 41.90 \\
        \textit{point.}-ORM & 79.76 & 20.20 & 33.43 / 45.42 & 41.73 / 49.79 \\
        \textit{point.}-ORM + SC & 79.83 & 23.80 & 34.80 / 44.45 & 41.70 / 49.78 \\
        \bottomrule
    \end{tabular}
    \end{adjustbox}
    \captionof{table}{Performance evaluations of different class of ORMs on $N=20$ sample output generations from Llama-3 8B SFT $1^{st}$ iteration. SC denotes Self-Consistency. RM only entries indicate Best-of-N (BoN) sampling based results. Mean of metrics reported on 3 groups of sampling results.}
    \label{tab:different_orms}
\end{table}

While \textit{pairwise}-loss RM training is generally considered more effective than \textit{pointwise}-loss RMs~\cite{liu2025improvingvideogenerationhuman}, empirical evidence remains divided. For instance, even accounting for differences in annotation guidelines and human expectations, \citeauthor{liu2025improvingvideogenerationhuman}(\citeyear{liu2025improvingvideogenerationhuman}) found pairwise RM training superior, whereas \citeauthor{wang2024helpsteer2preference}(\citeyear{wang2024helpsteer2preference}) reported better results with pointwise RM training. Our study adds to this debate with empirical evidence showing \textit{pointwise}-ORM outperforming \textit{pairwise}-ORM, significantly so on the MATH and \texttt{SpaRP} datasets (Table~\ref{tab:different_orms}). Both models are trained on a balanced set of positive and negative instances based on final answer outcomes, with \textit{pairwise}-ORM forming pairs in reference to given contexts. Furthermore, \texttt{SPARE}-ORM also underperforms \textit{pointwise}-ORM, despite incorporating \textit{superior} pairs selected based on both outcome and mean aggregated reasoning scores.

\section{Use of AI Assistance}
\label{sec:ai_assistance}
We used generative AI tools exclusively for grammar and language refinement. All content was subsequently reviewed and revised by the author(s), assuming full responsibility for the final version of the manuscript and its submission.


\end{document}